\documentclass[manuscript, screen]{jair}
\usepackage[english]{babel}
\usepackage[utf8x]{inputenc}
\usepackage{amsmath}
\usepackage{amsfonts}
\usepackage{graphicx}
\usepackage{subcaption}
\usepackage{threeparttable}
\usepackage{booktabs}
\usepackage{subcaption}
\usepackage{url}
\usepackage[colorinlistoftodos]{todonotes}
\newcommand{\comm}[1]{}

\setcopyright{cc}
\acmDOI{10.1613/jair.1.xxxxx}

\JAIRAE{Insert JAIR AE Name}
\JAIRTrack{} 
\acmVolume{4}
\acmArticle{6}
\acmMonth{10}
\acmYear{2026}

\RequirePackage[
  datamodel=acmdatamodel,
  style=acmauthoryear,
  backend=biber,
  giveninits=true,
  uniquename=init
  ]{biblatex}

\begin{document}

\title[Predicate Neuron]{Predicate Neuron: Differentiable Quantization of Predicate Relations in Gradient-based Learning}

\author{Ioannis Konstantoulas}
\authornote{Corresponding Author.}
\orcid{0009-0002-8438-6010}
\email{konstantou@ceid.upatras.gr}
\affiliation{%
  \institution{Department of Electrical and Computer Engineering, University of Patras}
  \city{Patras}
  \country{Greece}
}

\author{Iliana Loi}
\orcid{0000-0001-9112-0638}
\email{loi@ceid.upatras.gr}
\affiliation{%
  \institution{Department of Electrical and Computer Engineering, University of Patras}
  \city{Patras}
  \country{Greece}}

\author{Kyriakos Sgarbas}
\orcid{0000-0002-1797-1343}
\email{sgarbas@upatras.gr}
\affiliation{%
  \institution{Department of Electrical and Computer Engineering, University of Patras}
  \city{Patras}
  \country{Greece}
}

\renewcommand{\shortauthors}{Konstantoulas, Loi \& Sgarbas}


\begin{abstract}

{\bf Background:}
    In machine learning, learned symbolic approaches model reasoning by extracting data patterns that can be expressed as formal logic rules or by imposing weight-based constraints that guide neurons to approximate logical relationships. However, these approaches typically rely on static architectures that prevent them from being widely incorporated into deep neural models, while hindering seamless compatibility with gradient-based optimization across various tasks.

{\bf Objectives:}
    To address the aforementioned problem, we introduce the Native Predicate Neuron, a differentiable logic reasoning neural module that enriches typical neural architectures with logic reasoning capabilities, in order to enhance their accuracy and convergence utilizing first-order logic quantization.

{\bf Methods:}
    The proposed neuron natively expresses logic reasoning between relations by learning to alternate between logic operations, leveraging a single operation-selection trainable parameter. We devise a tailored gradient mechanism, where the forward pass operates through a dual-nature formulation that performs hard logical operations during training while maintaining differentiability for gradient propagation during backpropagation.
Moreover, the Predicate Neuron enables the approximation of both continuous and non-continuous functions, by employing a quantization method to quantize both the neuron's input and choice state, facilitating convergence while explicitly resolving discontinuities that standard deep learning architectures struggle with.

{\bf Results:}
    The Predicate Neuron is trained and evaluated across various deep learning architectures, including Fully-Connected, Convolutional, Recurrent and Transformer-based models, and benchmark datasets such as MNIST and Wikitext. Our results show that incorporating Predicate Neuron modules into neural structures accelerates convergence, while improving testing accuracy and accordingly reducing loss in the general case. In niche cases, where discontinuities or monotonic reasoning is required, the improvement in both convergence, accuracy and loss are staggering.

{\bf Conclusions:}
    Overall, the proposed neuron can be natively incorporated into typical neural architectures, enabling fuzzy logic operations without decreasing the pre-existing neural network performance.
    The Predicate Neuron and respective custom layers integrating it, are available in Keras and Pytorch libraries \footnote{\url{https://github.com/IoannisKonstantoulas/predicateneuron}} \footnote{The Predicate Neuron is also available as a Python package and can be downloaded from: \url{https://pypi.org/project/predicateneuron}}, facilitating its integration with a plethora of deep learning models.

\end{abstract}




\maketitle

\section{Introduction}

In recent years, researchers are turning to Artificial Intelligence (AI) to either provide solutions to tasks stemming from a plethora of scientific areas - from engineering to medicine and social sciences, as well as everyday tasks - or develop optimized models in terms of inference times and usage of fewer computational resources. Even though the premise of neural networks is to mimic the way the neurons of the brain work, they struggle with logical reasoning \cite{Bhuyan2024neuro}.

Neurosymbolic frameworks \cite{Nawaz2025review}, typically deep neural networks that incorporate monotonic reasoning and stochastic machine learning, aim to cover this gap. Prominent approaches \cite{Mommen2026fully, Petersen2022deep, Evans2018learning, Sourek2018lifted} are designed to explicitly model rule-based reasoning within their custom neurons, or constrain their weights to force them to act as logical gates, ultimately converging on the underlying logic relationships of a given problem. Such systems can effectively approximate formal logic reasoning, but reliably converging to solid and effective logical structures still remains a challenge, with recent Neurosymbolic networks \cite{Tran2026Reasoning} employ energy-based constraints or continuity-enforcing strategies \cite{Riegel2020} to facilitate convergence. However, the incorporation of first order logic rules or operations requires specific neural architectural design choices to be made, hindering the ability of such modules to be migrated to heterogeneous gradient-based structures, while the capability of such models to be used for a variety of tasks is limited.

Furthermore, the gradient-based learning procedure leveraged by neural network prevents them from learning discontinuities natively, with most works resorting to model discontinuities in a continuous manner e.g. as probability distributions \cite{Petersen2022deep}, continuous mathematical functions \cite{Badreddine2022}, etc. Solutions to this issue consist to utilize reasoning modules to express discontinuities in a quantized-like way within the model, so as to use it as an a priori knowledge to learn the function describing the to-be-investigated problem, without altering its non-continuous nature.

In this work, we present the Native Predicate Neuron, a native differentiable neuron/layer that can be seamlessly integrated into common deep learning architectures. The motivation behind this work originally was to narrow the gap between symbolic approaches and neural networks incorporating the rigid logical structure of symbolic approaches into the differentiable approximating nature of deep neural structures, to facilitate their convergence and overall performance.
The Native Predicate Neuron can express logic reasoning natively within the neural network during training and inference, maximizing its reasoning capacity. We model the predication of concepts natively within the neural network by leveraging a quantization technique to effectively guide the model to approximate both continuous and non-continuous operations, without the need for explicit handling of discontinuities.
The Native Predicate Neuron is available as Keras \cite{Chollet2015keras} and Pytorch \cite{Paszke2019pytorch} libraries, providing easily integratable built-in modules for deep learning architectures \footnote{\url{https://github.com/IoannisKonstantoulas/predicateneuron}} \footnote{\url{https://pypi.org/project/predicateneuron}}.

Our contribution is two-fold as described below:

\begin{itemize}
    \item Providing a differentiable logic reasoning neural module, named the Native Predicate Neuron, which can be integrated natively into any typical neural architecture to enrich it with reasoning.
    \item
    We also introduce a predicate relation quantization schema that can be appended to common deep learning models. This method can capture and express discrete atomic knowledge while also enabling the neural network to exploit this knowledge for expressing mathematical discontinuities that common neural architectures cannot.
    \end{itemize}

\section{Related Work}

Prominent works in Neurosymbolic AI providing solutions towards logic reasoning are neural networks that incorporate logical rules within their cost function or structure as soft constraints to guide the learning process of the model. Characteristic examples include the Logical Neural Networks (LNNs) \cite{Riegel2020}, which leverage weight constraints to force their neurons to act as logical gates based on the hypothesis that each neuron corresponds to a predicate atom of formal logic (i.e., a meaning or a logical operation). Each neuron in an LNN returns a pair of values in $[0, 1]$ as lower and upper bounds instead of single-point values to handle unknown and contradictory knowledge alongside true/false statements. Similar models include the Neural Logic Machines (NLMs) \cite{Dong2019} and the Logic-Integrated Neural Networks (LINN) \cite{Shi2020}, which utilize neural networks to approximate basic logical operations, and logic predicates are represented as tensors. Additionally, in NLMs \cite{Dong2019}, their weights correspond to logic quantifiers, while LINNs \cite{Shi2020} dynamically adjust their neural architecture so as to provide a solution to the input logical expression. It is worth noting DeepProbLog \cite{Manhaeve2018deepproblog}, one of the first frameworks in this field, which entangles standard deep neural networks with probabilistic-logical modeling and reasoning via an extension of the ProbLog \cite{DeRaedt2016statistical} probabilistic programming language.

More works combining symbolic reasoning with gradient-based neural learning are the Logic Tensor Networks (LTNs) \cite{Badreddine2022} and their extension, the Structural Logic Tensor Networks (sLTN) \cite{Rinaldi2026sltn}. Just as in the above cases \cite{Riegel2020, Shi2020, Dong2019}, LTNs \cite{Badreddine2022} embed logic reasoning directly into the computational graph and the optimization objective of the model (i.e., as soft constraints, penalizing the network during training if it violates explicit domain knowledge), in the form of the first-order (FOL) logic language Real Logic. Real Logic is utilized to describe and estimate common deep learning tasks such as multi-label classification, regression, query answering, etc., in which language logical operations are approximated using continuous mathematical functions from fuzzy logic semantics (e.g., t-norm, t-conorm, fuzzy implication, etc.) and logic quantifiers are defined with the use of aggregation. The work in \cite{Skryagin2023scalable} introduces also a Deep Probabilistic Programming Language (DPPL) that uses explicit Neural-Probabilistic Predicates and map probabilistic predicate relations directly into differentiable GPU tensor operations to accelerate reasoning within the neural network, without hindering performance. In the more recent work \cite{Rinaldi2026sltn}, Structural Logic Tensor Networks encode structural information as components of Real Logic to formulate temporal, sequential, and relational constraints between input data, unlike LTNs that are designed to process data that contain uncorrelated, unorganized samples.

Another form of Neurosymbolic AI is the Logic Gate Networks (LGNs), which explicitly model their neurons as logical gates, producing fast architectures composed entirely of logic operations, in contrast with the models that fall under the category of Logical Neural Networks that exploit logical gates in the context of symbolic reasoning. In \cite{Petersen2022deep}, Differentiable Logic Gate Networks are proposed that circumvent the non-trainable nature of fixed logic‑gate architectures and enable gradient-based optimization. To that end, a probability distribution over possible gates is learnt for each neuron, followed by a discretization process to select which logic gate will represent which neuron (based on the highest probability) and ultimately provide a logic gate network. Convolutional \cite{Petersen2024convolutional} and Recurrent \cite{Buhrer2025} Differentiable Logic Gate Networks are also proposed in  more recent literature. In the former work \cite{Petersen2024convolutional}, LGNs are extended with neurons using logic gate trees akin to convolutional structures, rather than comprising of a single gate, for effectively learning spatial dependencies and, thus, increasing testing accuracy in multi-label classification tasks, while offering faster training and inference times. Recurrent Deep Differentiable Logic Gate Networks (RDDLGN) \cite{Buhrer2025} introduce shared‑weight recurrent mechanisms that provide memory, enabling LGNs to capture temporal dependencies and learn sequence‑to‑sequence tasks. The most recent work in this field \cite{Fojcik2026lilogic}, LILogic Net leverages a connectivity strategy to enable network connectome sparsity so as to significantly reduce the number of logic gates utilized in the model and ultimately achieve better performance compared with previous larger in magnitude LGNs \cite{Petersen2022deep, Petersen2024convolutional}.

Nevertheless, LGNs, aiming to provide a neural architecture completely comprised of logic gates to minimize the computational cost and enable fast inference, do not offer universally integratable modules, but rather require to follow specific architectures to resemble the functionality of more complex deep learning models. Moreover, LNN- and LTN-like architectures explicitly model symbolic reasoning rules within their computational graph enabling differential learning, yet are limited by the logic rules embedded in their architecture not being able to describe any deep learning task. The Native Predicate Neuron addresses the above shortcomings identified in the aforementioned works, since it can be integrated into any deep learning architecture, enhancing it with logic reasoning capabilities, ultimately used to solve multiple deep learning tasks, from FOL problems, to multi-label classification and mathematical problems, with faster convergence and better testing accuracy than its common neural network counterparts.

\section{Methodology}

\subsection{Native Predicate Neuron}

The proposed module implements a logic operation neuron that "selects" one of three fuzzy logic-set operations, namely intersection, union, and symmetric difference, to be employed. The neuron is governed by a single learnable parameter, $\theta$, which determines the choice of the logical function. To enable end‑to‑end training, we introduce a tailored gradient mechanism that employs a smooth gradient with respect to the operation‑selection parameter, thereby ensuring differentiability during backpropagation.

We use the Predicate Neuron assuming that a value $1$ indicates that a predicate is true, whereas the $0$ the value denotes that it is not known whether the predicate is true or false. To define a predicate $\phi$ as false, a second predicate is supplied or learned by the quantizing network preceding it, which predicate is the negation $\phi'$ of $\phi$. In neural networks that use the Sigmoid for predicate quantization as described later (see Subsection \ref{sec:non_continuous_experiments}), this is simply done by multiplying any conditional, before feeding it to the the Sigmoid, by $-1$, and supplying it to a different distinct Sigmoid activator. The concept of implication is handled natively by the fact that the output of neurons is fed as input to others, in this sense we can say that $\phi = true$ and $\psi = true$ implies $\phi \land \psi$. If we supply one of the inputs of a union neuron with the value $1$, then $\phi = true$ implies $\phi \land true$. Outside the Predicate Neurons, one way to affect the values of the neural network is to multiply the values of the predicates with the outputs of neurons.

The Predicate Neuron has two modalities, namely the hard and soft forward pass, that can dynamically change during training through the use of a hyperparameter, annotated as \textit{hard\_forward}. In the hard forward pass, the neuron performs a discrete selection of a single logical operation to produce its output, while the soft forward pass computes a per-operation weighted aggregation of the three core logic-set operations between the two inputs. The soft forward pass is better used at the beginning of training to improve early stage acquisition of logic relations, while the hard forward pass presents more stable outputs helping the optimizer train the rest of the network. Furthermore, we introduce a \textit{hard\_backward} hyperparameter that controls whether input gradients ($\frac{\partial L}{\partial a}, \frac{\partial L}{\partial b}$) are derived from the selected logical operation (hard forward pass) or propagated through the differentiable soft formulation (soft forward pass). The mathematical formulation of the forward and backward pass as well as details regarding the hyperparameters governing the proposed neuron, are provided in the subsequent subsections.

\subsubsection{Mathematical Formulation}
\label{sec:math_formulation}

\paragraph{Hard Forward Pass:}

The neuron is fed with two inputs, $a$ and $b$ in the range $[0, 1]$, and produces an output corresponding to one of the following operations as defined in Eq. \ref{eq:logic_op_functions}:

\begin{equation}
y_h =
\begin{cases}
ab & \theta < 0.25 \\
a + b - ab & 0.25 \le \theta < 0.75 \\
a + b - 2ab & \theta \geq 0.75
\end{cases}
\label{eq:logic_op_functions}
\end{equation}

In particular, $y_{\cap} = ab$ is the product t-norm representing intersection, $y_{\cup} = a + b - ab$ is the algebraic sum t-conorm corresponding to the union operation, and $y_{\triangle} = a + b - 2ab$ describes the algebraic symmetrical difference. Therefore, the forward pass produces a hard and piecewise constant with respect to $\theta$.

\paragraph{Soft Forward Pass:}

The neuron is fed with two inputs, $a$ and $b$ in the range $[0, 1]$, and produces an output corresponding to the per-operation weighted aggregate as defined in Eq. \ref{eq:soft_pass}:

\begin{equation}
    y_s = p_0ab + p_1(a + b - ab) + p_2(a + b - 2ab)
    \label{eq:soft_pass}
\end{equation}

\paragraph{Backward Pass:}

Following the classical backpropagation algorithm, we compute the gradient defined in Eq. \ref{eq:dy}:

\begin{equation}
    \mathrm{dy}=\frac{\partial L}{\partial \hat{y}}
    \label{eq:dy}
\end{equation}

where $L$ is the computed loss over the whole neural structure, and $\hat{y}$ is the output of the Predicate Neuron formulated as (Eq. \ref{eq:y_hat}):

\begin{equation}
    \hat{y} = p_0ab + p_1(a + b - ab) + p_2(a + b - 2ab)
    \label{eq:y_hat}
\end{equation}

where $p_i$ are the probabilities yielded by the softmax computation (Eq. \ref{eq:p_i}):

\begin{equation}
   p_i = \frac{e^{z_i}}
{e^{z_0}+e^{z_1}+e^{z_2}}
\label{eq:p_i}
\end{equation}

with $z_i$ specified as:

\begin{equation}
    z_i(\theta) = \frac{-(\theta - a_i)^2}{T}
    \label{eq:z_i}
\end{equation}

where $a_i$ is an auxiliary parameter with values $\{0, 0.5, 1\}$, referred to as "position", which facilitates the computation of the gradient of $\theta$. In particular, the three logical operations are positioned on the $\theta$-axis as dictated below:

\begin{equation}
    a_0=0,\qquad a_1=0.5,\qquad a_2=1.
\end{equation}

with $a_0$ pointing to intersection, $a_1$ corresponding to union, and $a_2$ representing symmetric difference. It is worth noting that the assignment of each position is based on their semantic distance, so that the neuron would seamlessly learn the switch from one logical operation to the other. For example, union is more "close" to the intersection operation (i.e. their in-between distance is translated as $0.5$), thus, less weight ($\theta$) updates need to be made for the neuron to converge to a valid prediction. To be accurate, if semantic distance between the logical operations is evaluated by truth table Hamming distance over $\{0, 1\}^2$, the OR operation has the closest distance between both AND and XOR, thus, we opt to position the OR selection area in between the two others.

\begin{figure}[htbp]
    \centering
    \begin{subfigure}[b]{0.32\textwidth}
        \centering
       \includegraphics[width=\textwidth]{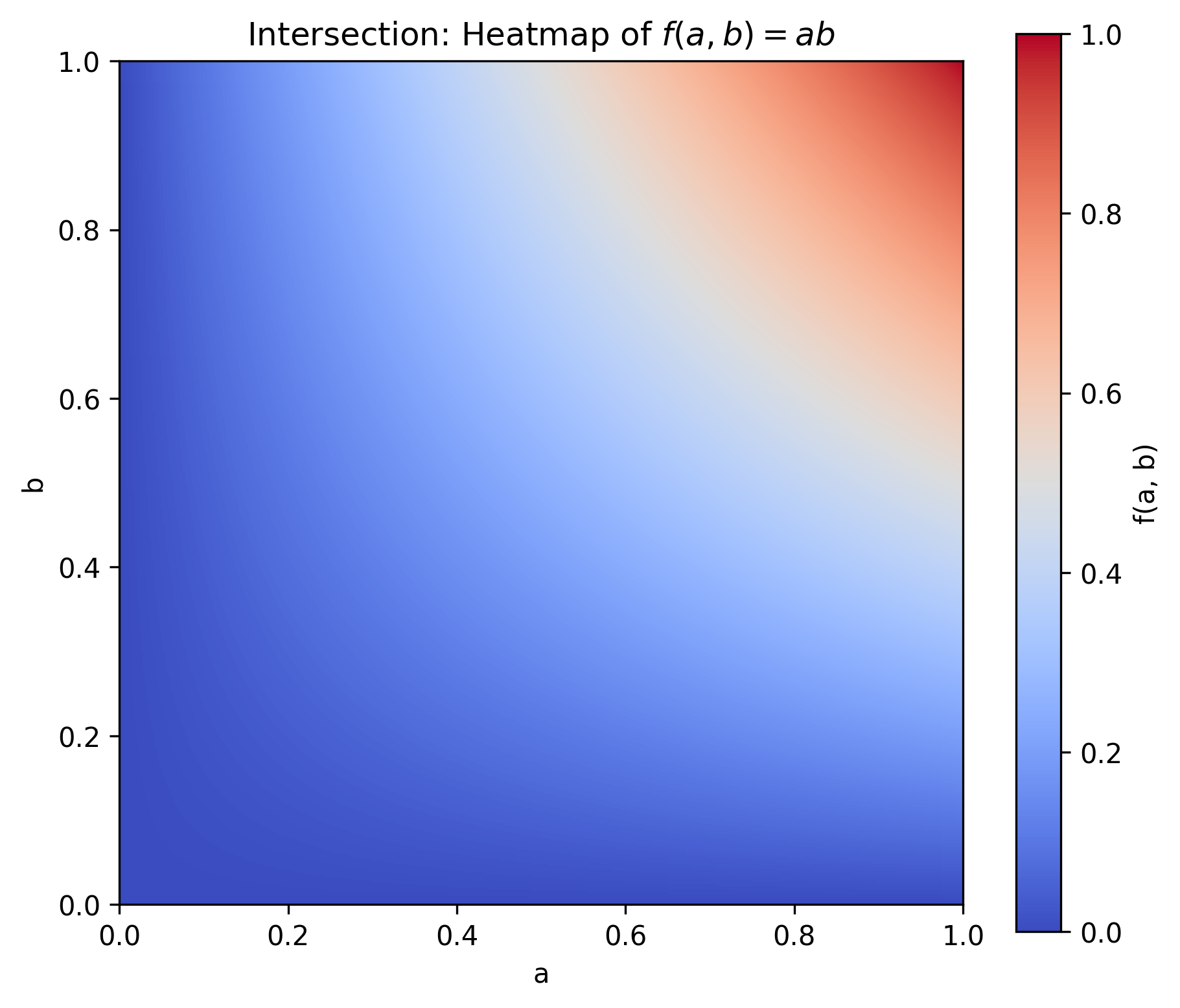}
    \end{subfigure} \hfill
    \begin{subfigure}[b]{0.32\textwidth}
        \centering
        \includegraphics[width=\textwidth]{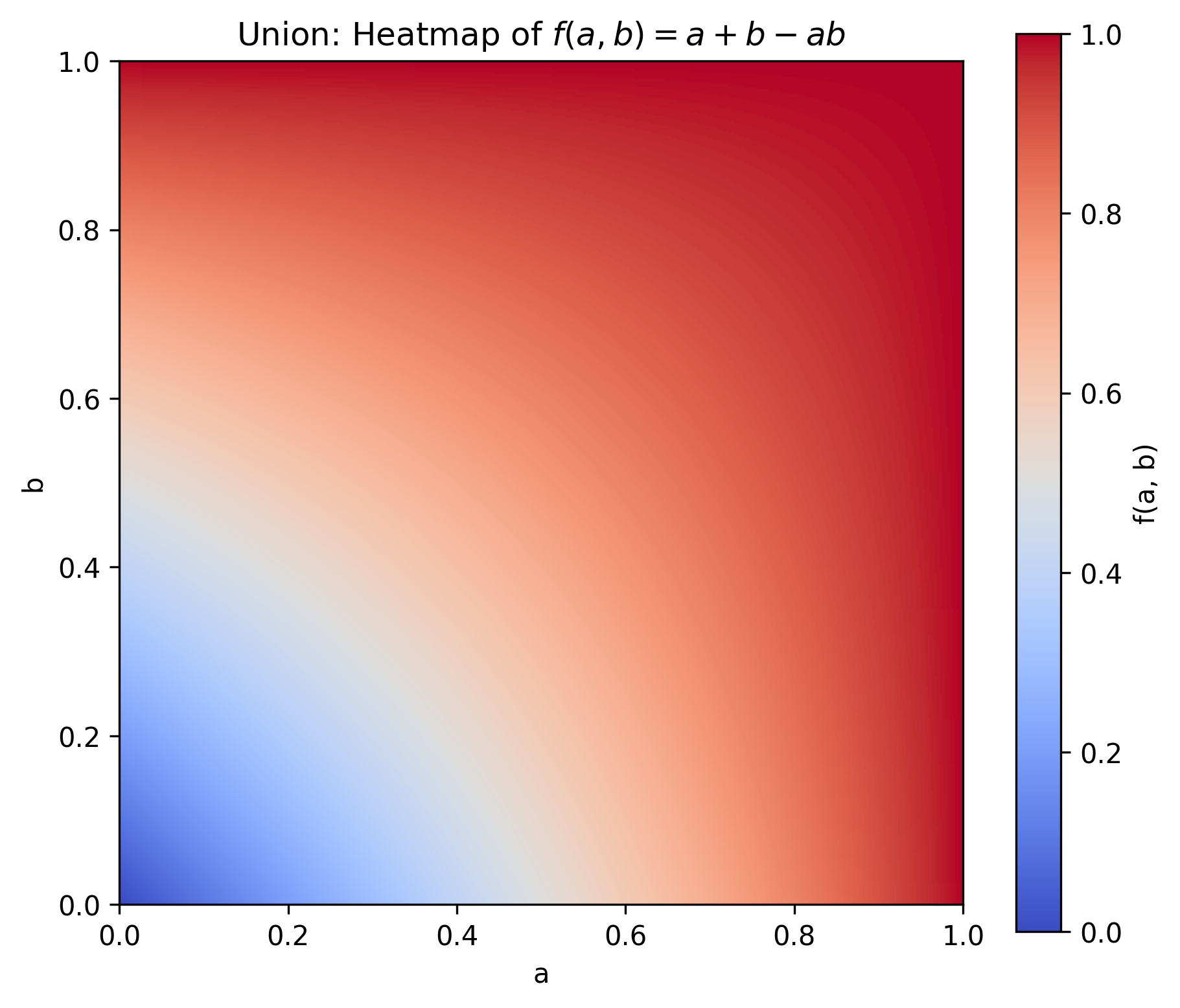}
    \end{subfigure} \hfill
    \begin{subfigure}[b]{0.32\textwidth}
        \centering
        \includegraphics[width=\textwidth]{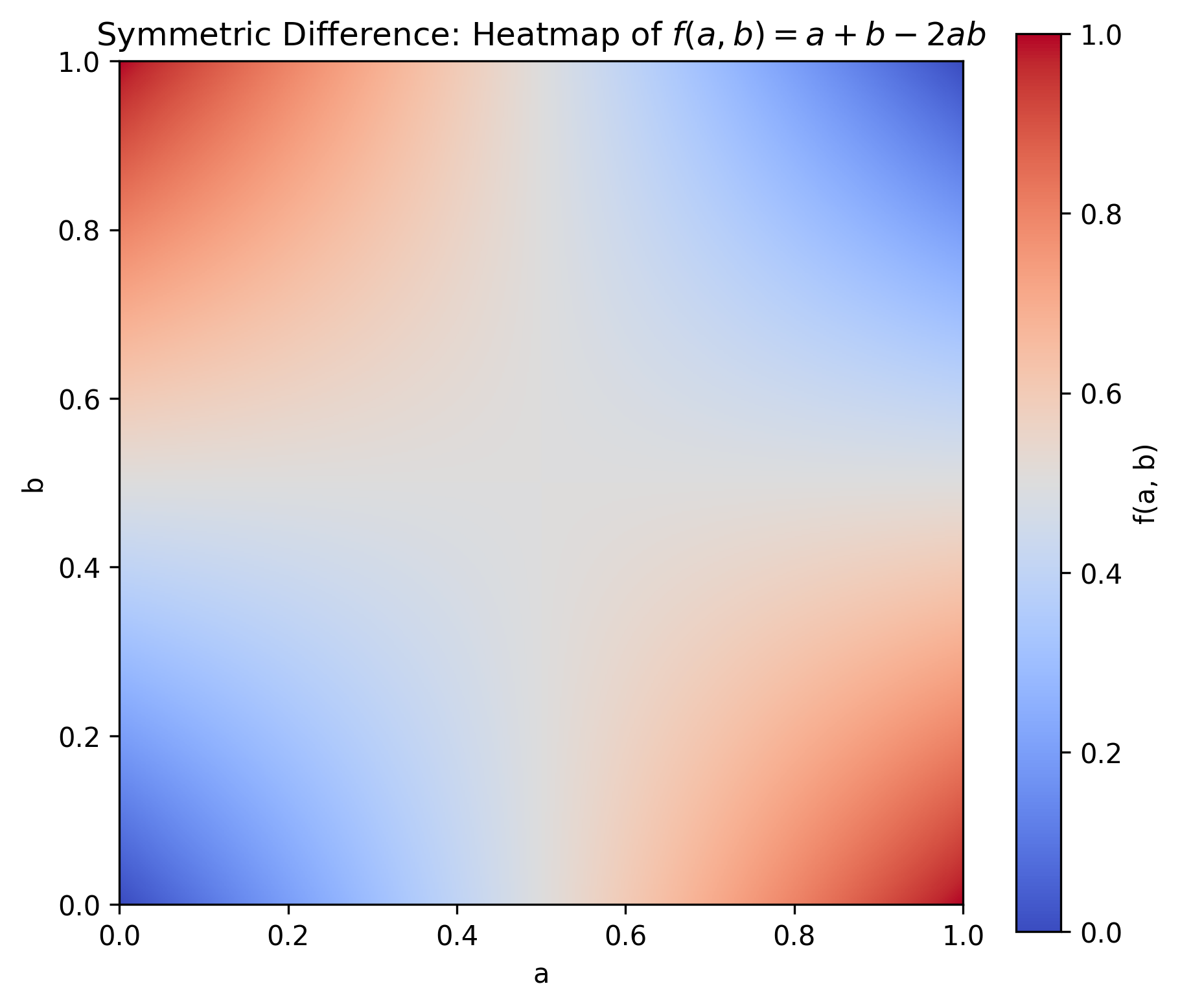}
    \end{subfigure}
    \caption{Heatmaps of the three logical operations, namely AND (left), OR (middle), and XOR (right).}
    \label{fig:logic_gate_functions}
\end{figure}

The distinction between these three logical operations is evident in Fig. \ref{fig:logic_gate_functions}, where their resulting values based on the inputs $a$ and $b$ are illustrated as a heatmap. Overall, $p_i$ is largest for the operation whose assigned position is closest to $\theta$. For example, if $\theta=0.1$, then $p_0$ is the largest value because $\theta$ is closest to $a_0$.

Following the backward pass, the gradients of $z_i$ are computed as follows (Eq. \ref{eq:dz_i}):

\begin{equation}
   \frac{\partial z_i}{\partial\theta} = -\frac{2(\theta-a_i)}{T}
\label{eq:dz_i}
\end{equation}

revealing that if $\theta$ is below a position $a_i$, then $dz_i/d\theta>0$, if $\theta$ is above $a_i$, then $dz_i/d\theta<0$, and, consequently, the greater distance between $\theta$ and $a_i$, the bigger the magnitude.

Then, the gradients of probabilities $p_i$ are calculated as dictated by Eq. \ref{eq:dp_i}:

\begin{equation}
\frac{\partial p_i}{\partial\theta} = p_i \left(\frac{\partial z_i}{\partial\theta} - \mathbb{E}_p\left[\frac{\partial z}{\partial\theta}\right]\right).
\label{eq:dp_i}
\end{equation}

where $\mathbb{E}_p\left[\frac{\partial z}{\partial\theta}\right]$ is the weighted average representing the expected derivative as defined in Eq. \ref{eq:weighted_average}:

\begin{equation}
    \mathbb{E}_p\left[\frac{\partial z}{\partial\theta}\right] = \sum_jp_j\frac{\partial z_j}{\partial\theta}
\label{eq:weighted_average}
\end{equation}

Given Eq. \ref{eq:weighted_average}, the Eq. \ref{eq:dp_i} will be written as:

\begin{equation}
    \frac{\partial p_i}{\partial\theta}
=
p_i
\left(
\frac{\partial z_i}{\partial\theta} - \sum_j p_j\frac{\partial z_j}{\partial\theta}\right).
\end{equation}

Subtracting the expected derivative is essential for maintaining the probabilities property to sum to $1$. So when one operation’s probability increases, at least one of the other operation’s probabilities must decrease.

Given the formulation in Eq. \ref{eq:y_hat}, we establish a smooth gradient with respect to the $\theta$ parameter, which we formulate as follows (Eq. \ref{eq:surrogate_gradient}):

\begin{equation}
    \frac{\partial\hat{y}}{\partial\theta}
=
\sum_i
\frac{\partial p_i}{\partial\theta}y_i =
\frac{\partial p_0}{\partial\theta}y_{\cap}
+
\frac{\partial p_1}{\partial\theta}y_{\cup}
+
\frac{\partial p_2}{\partial\theta}y_{\triangle}
\label{eq:surrogate_gradient}
\end{equation}

If training the parameter $\theta$ with hard forward pass is desired, we opted to utilize this surrogate gradient since the gradient of the hard output of the forward pass ($y$), is $\frac{\partial y}{\partial\theta}=0$, hindering the learning of the $\theta$ parameter.

Overall, the backward process requires computing the following gradients (Eq. \ref{eq:custom_gradient}):

\begin{equation}
    \frac{\partial L}{\partial a},
    \qquad
    \frac{\partial L}{\partial b},
    \qquad
    \frac{\partial L}{\partial\theta}
    \label{eq:custom_gradient}
\end{equation}

By applying the chain rule for $\theta$, since $\mathrm{dy}=\frac{\partial L}{\partial \hat{y}}$, then:

\[
\frac{\partial L}{\partial\theta}
=
\frac{\partial L}{\partial \hat{y}}
\frac{\partial \hat{y}}{\partial\theta}.
\]

which for tensors or batches it becomes:

\[
\frac{\partial L}{\partial\theta}
=
\sum_k
\mathrm{dy}_k
\frac{\partial \hat{y}_k}{\partial\theta}.
\]

If $\theta$ is a scalar shared across a batch, this sums the gradient contribution from every example.

Moreover, the gradients with respect to the inputs $a$ and $b$ are:
For intersection:

\[
\frac{\partial y_{\cap}}{\partial a}=b,
\qquad
\frac{\partial y_{\cap}}{\partial b}=a.
\]

For union:

\[
\frac{\partial y_{\cup}}{\partial a}=1-b,
\qquad
\frac{\partial y_{\cup}}{\partial b}=1-a.
\]

And for symmetric difference:

\[
\frac{\partial y_{\triangle}}{\partial a}=1-2b,
\qquad
\frac{\partial y_{\triangle}}{\partial b}=1-2a.
\]

which are then used to compute the gradients $\frac{\partial L}{\partial a}$ and $\frac{\partial L}{\partial b}$ as follows:

\[
\frac{\partial L}{\partial a}
=
\frac{\partial L}{\partial \hat{y}}
\frac{\partial \hat{y}}{\partial a}
\]

and

\[
\frac{\partial L}{\partial b}
=
\frac{\partial L}{\partial \hat{y}}
\frac{\partial \hat{y}}{\partial b}.
\]

So the gradients with respect to the input variables are the exact derivatives of whichever operation was selected in the forward pass. In more detail:

\[
\frac{\partial \hat{y}}{\partial a}
=
\begin{cases}
b,&g=0\\
1-b,&g=1\\
1-2b,&g=2
\end{cases}
\]

Similarly,

\[
\frac{\partial \hat{y}}{\partial b}
=
\begin{cases}
a,&g=0\\
1-a,&g=1\\
1-2a,&g=2.
\end{cases}
\]

with $g$ being the set-operation selection, i.e. $0$ for intersection, $1$ for union, and $2$ for symmetric difference. The above formulation corresponds to the hard backward pass, meaning that the input derivatives are routed strictly through the selected logical operation $g \in \{0, 1, 2\}$. On the contrary, depending on the \textit{hard\_backward} hyperparameter, the input gradients may also flow through the soft weighted aggregation as dictated by Equations \ref{eq:soft_backward_pass_1} and \ref{eq:soft_backward_pass_2}:

\begin{equation}
	\frac{\partial \hat{y}}{\partial a} = p_0 b + p_1 (1-b) + p_2 (1-2b)
	\label{eq:soft_backward_pass_1}
\end{equation}

\begin{equation}
	\frac{\partial \hat{y}}{\partial b} = p_0 a + p_1 (1-a) + p_2 (1-2a)
	\label{eq:soft_backward_pass_2}
\end{equation}

Overall, it is worth noting that regardless of how the input gradients are computed, the computation of the gradient with respect to $\theta$ (i.e. $\frac{\partial L}{\partial\theta}$) is always facilitated by the smooth gradient described in Eq. \ref{eq:surrogate_gradient} guaranteeing differentiability in the backwards pass.

\subsubsection{Hyperparameters}

Apart from the number of epochs and learning rate, we define three primary hyperparameters for the Predicate Neuron:
\begin{enumerate}
\item \textbf{Temperature ($T$)}: Controls how sharply the soft probabilities $p_i$ (Equations \ref{eq:y_hat} and \ref{eq:p_i}) respond to change in $\theta$,
\item \textbf{Hard Forward (\textit{hard\_forward}) Hyperparameter}: A Boolean flag  that enables the switch between the hard and forward pass,
\item \textbf{Hard Backward (\textit{hard\_backward}) Hyperparameter}: A boolean flag controlling whether the gradients of the loss w.r.t the input variables (i.e. $\frac{\partial L}{\partial a}$,  $\frac{\partial L}{\partial b}$) are routed via the selected discrete logical operation or the soft formulation defined in Eq. \ref{eq:y_hat}.
\end{enumerate}

During training, the \textit{hard\_forward} hyperparameter can be additionally enabled with a similar effect of temperature decrease resulting in a discrete selection of set operation. In cases where the Predicate Neuron is a part of a deep neural network and it reaches a point of minimal local improvement during optimization (has settled to a set of logical operations), the $\theta$ weight can also be designated as non-trainable allowing for a native transformation to a rigid fuzzy logic network. In such a case, the Predicate Neuron operates as an embedded fuzzy logic structure within the network, ultimately accelerating training, since the total number of trainable parameters is reduced (only the layers preceding and succeeding our neuron, can resume their training). This parameter-frozen logic structure can be utilized in meta-learning to initialize a new network with the same logic structure.

\begin{figure}[h!]
    \centering
    \includegraphics[width=\linewidth]{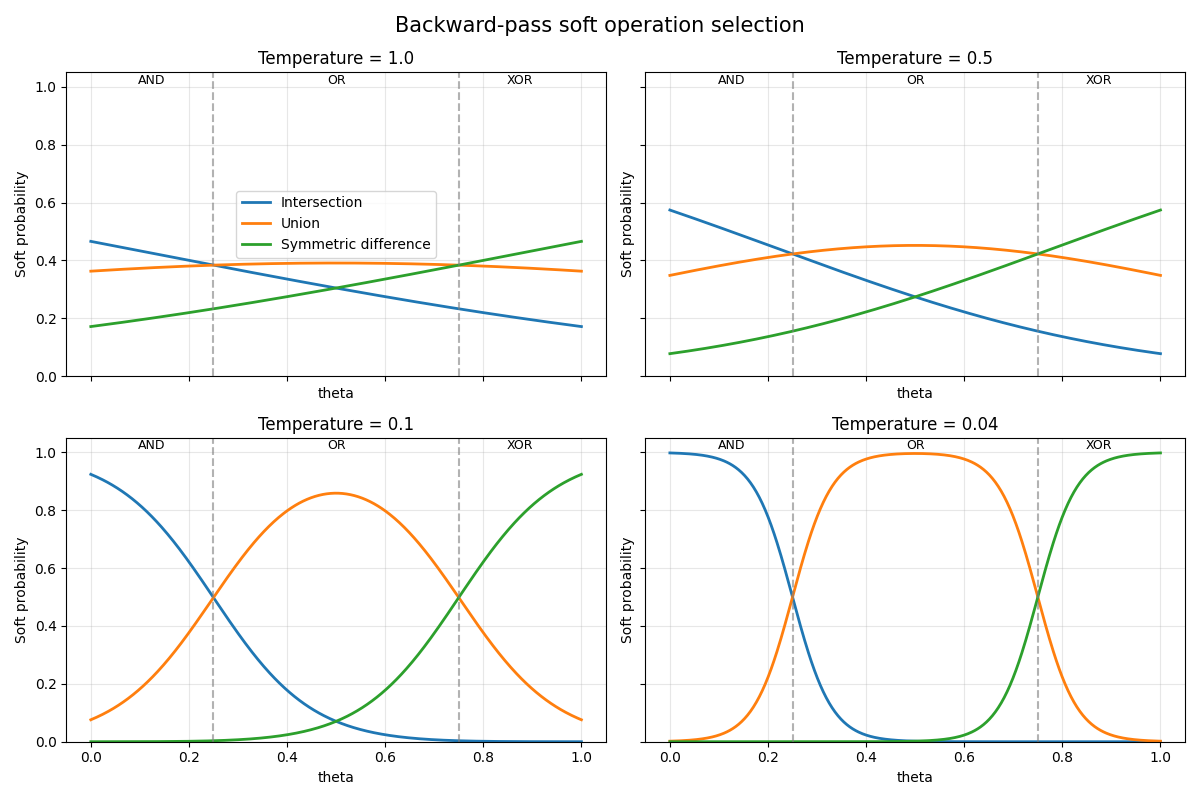}
    \caption{A visual representation on how the Predicate Neuron switches between the three logical operations based on the value of the temperature hyperparameter.}
    \label{fig:temperature_analysis}
\end{figure}

As for the temperature hyperparameter $T$, its value should match the one of the learning rate, with values towards $0.1$ facilitating convergence to an operation choice, as shown in Fig. \ref{fig:temperature_analysis}.

\section{Results}

In order to evaluate the effectiveness of the Native Predicate Neuron we conducted a set of experiments measuring performance in terms of accuracy, loss and convergence, across different datasets and architectures in which the neuron was integrated. The latter models incorporating the Predicate Neuron will be hereafter referred to as \textit{Logic Networks}, and are compared against their neural network counterparts, i.e., standard model architectures without the proposed neuron. All results outlined in the following Subsections, correspond to the mean values obtained over $10$ independent runs for each experiment, except for RNN, which we evaluated over $3$ runs per experiment due to its very long training times.

The experiments were conducted on a python keras running on an AMD Ryzen 8 5700X 8-Core Processor at 4.5 GHz Processor.

\subsection{Fundamental Neuron Benchmarking}
\label{sec:fund_benchmarking}

The following $4$ experiments aim to showcase the ability of the Predicate Neuron to enrich Fully-Connected architectures with reasoning capabilities so as to approximate more accurately both conventional logic reasoning problems and more complex ones, such as the Parity problem discussed in subsection \ref{sec:parity_experiment}, known for its difficulty for gradient-based learning.

\subsubsection{Fundamental Operations}
\label{sec:fund_ops}

In our first experiment, we feed the Predicate Neuron with $3$ predicates, $a$, $b$, and $c$, with each combination of inputs yielding one discrete logical operation. In particular, $<a, b>$ results in AND, $<a, c>$ in OR, and $<b, c>$ in XOR. Hence, we created a layer with $3$ Predicate Neurons.

\begin{figure}[h!]
    \centering
    \includegraphics[width=\linewidth]{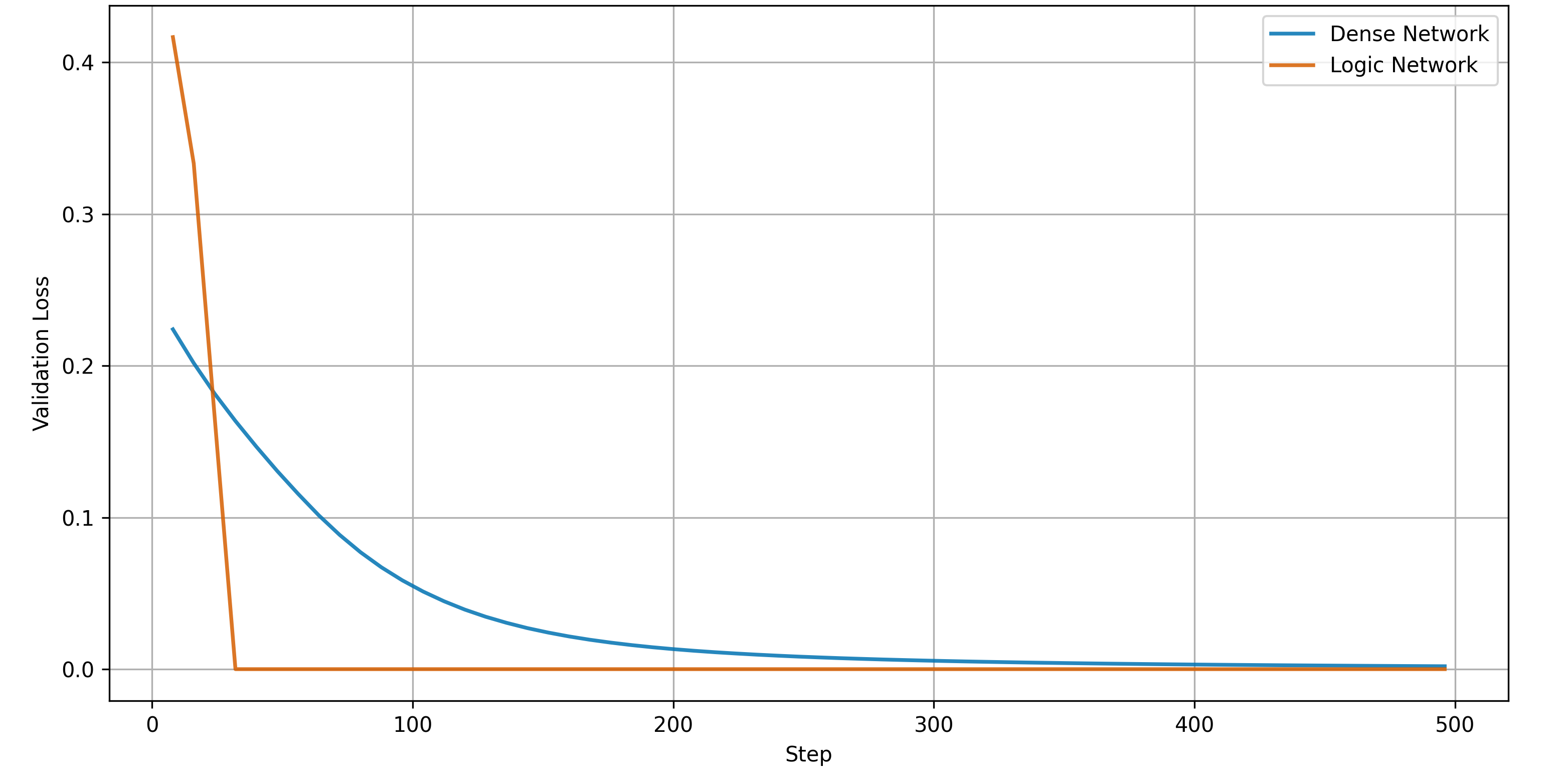}
    \caption{Validation Loss of the Predicate Neuron (orange graph) and the Dense network (blue graph).}
    \label{fig:and_or_xor_experiment}
\end{figure}

The main aim of this experiment is to assess whether our neuron is capable of converging to low-level logical operations, i.e., validating its basic functionality. We evaluate this configuration against a Fully-Connected Network in terms of Mean Squared Error Loss (MSE), as illustrated in Fig. \ref{fig:and_or_xor_experiment}. The Fully-Connected Network also comprises an input layer with $3$ inputs, $1$ hidden layer with $10$ units and ReLU activation, and an output layer with $3$ outputs, and Sigmoid activation, having a total of $33$ parameters.

The Dense network reports an $MSE = 0.0192$ on the validation set, while the Logic Neuron achieves an $MSE \sim= 0.000$, as can be also shown in Fig. \ref{fig:and_or_xor_experiment}. Reaching $0$ loss is reasonable, due to the network's hard forward pass and thus explicitly selecting a distinct logical operation. From Fig. \ref{fig:and_or_xor_experiment} we also observe that the Logic Network converges earlier, around step $32$, compared to the Dense network without Predicate Neurons, that never truly reaches zero loss. Apart from accuracy, in this test, the Predicate Neuron surpasses the Dense in terms of training speed, performing $50$ training steps in $1.07$ seconds, while the Dense network runs $50$ steps in $1.16$ seconds, both in CPU training.

As claimed earlier the Predicate Neuron naturally provides ease of explanation regarding the operations being done and by observing the values of the $\theta$ parameter, we may acquire an empirical sense of what the neural network learned. For this experiment in one of the experimental runs, the model results in the following values for the three Predicate Neurons; $0.1186$ which corresponds to the AND operation, $0.3615$ representing the OR operation, and $0.8517$ matching XOR. These values provide a perfect solution, because the \textit{hard\_forward} hyperparameter is enabled during training and reaches zero loss almost instantly.

\subsubsection{The 20-atom Experiment}
\label{sec:20_atom_experiment}

We extend our evaluation on a logic predicate dataset, describing the sex of a person and whether they are a parent (mother or father). The dataset consists of $20$ input predicate atoms, and $21$ output predicate atoms, where $20$ of those are the same atoms as the input and the last output is a satisfiability flag specifying whether the input is a stable model according to stable model semantics \cite{Fitting1992michael}. Another way to look at the satisfiability flag is that it is false if the input set of atoms implies a contradiction, or else it is true. For instance, a set of atoms reflecting a possible/real-world scenario, e.g., the following sentence "someone who is a mother is a woman" constitutes a satisfiable case, while "a person is a parent and not a parent" is a contradiction, therefore an unsatisfiable case. In the context of this experiment, we utilized a more complex architecture for the Logic Network, with $20$ inputs, $1$ Predicate layer with $190$ units, and $21$ outputs. The hidden layer of this model is an across input pairwise Fully-Connected layer, since it represents all pairwise combinations between the two inputs of each Predicate Neuron and the multi-variable input, i.e. $\binom{20}{2} = 190$. This layer is called \textit{Maximal Predicate Layer}.

Similar to the above test (i.e. subsection \ref{sec:fund_ops}), we compare our neuron with Fully-Connected Networks, one that has $1$ hidden layer with $10$ units ($441$ parameters), and one that has $3$ hidden layers with $40$ units each (overall $5k$ parameters). Binary Cross-entropy was used as the loss function in this case, with Dense model with $5k$ parameters reaching a loss of $BCE = 3.8348*10^{-6}$, simpler Dense exhibiting a loss of $BCE =  1.1609 * 10^{-04}$, while the network with the Predicate Neuron outperforms both Dense models with a loss of $BCE = 9.8694*10^{-7}$. Furthermore, the $5k$ Fully-Connected network requires $73.27$ seconds for $10000$ steps of training on the CPU, and, on the contrary, the Logic Network performs $10000$ steps in $82.26$ seconds on CPU training.

\begin{figure}[htbp]
    \centering
    \begin{subfigure}[b]{0.90\textwidth}
        \centering
       \includegraphics[width=\textwidth]{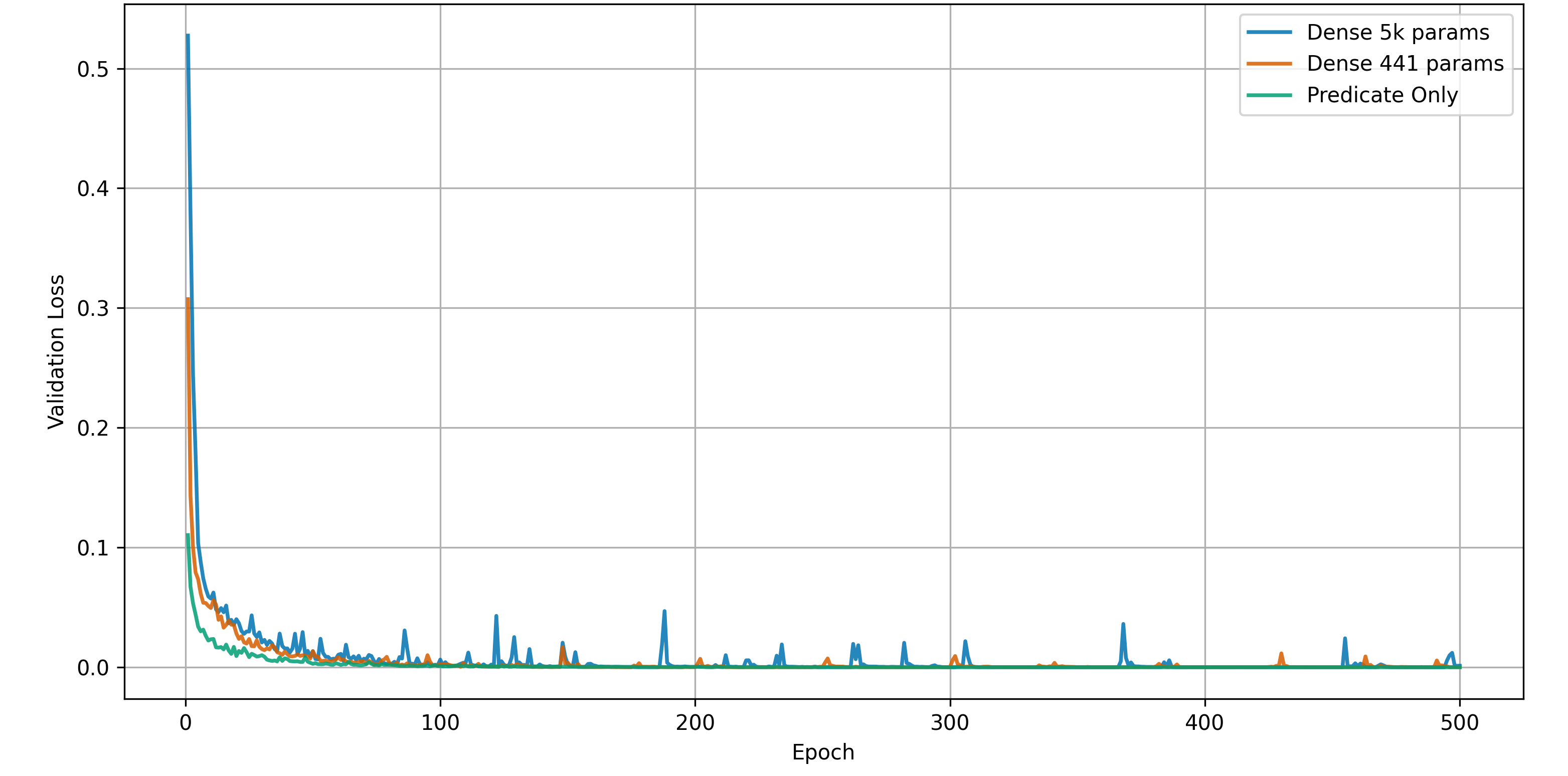}
       \caption{}
    \end{subfigure} \hfill
    \begin{subfigure}[b]{0.90\textwidth}
        \centering
        \includegraphics[width=\textwidth]{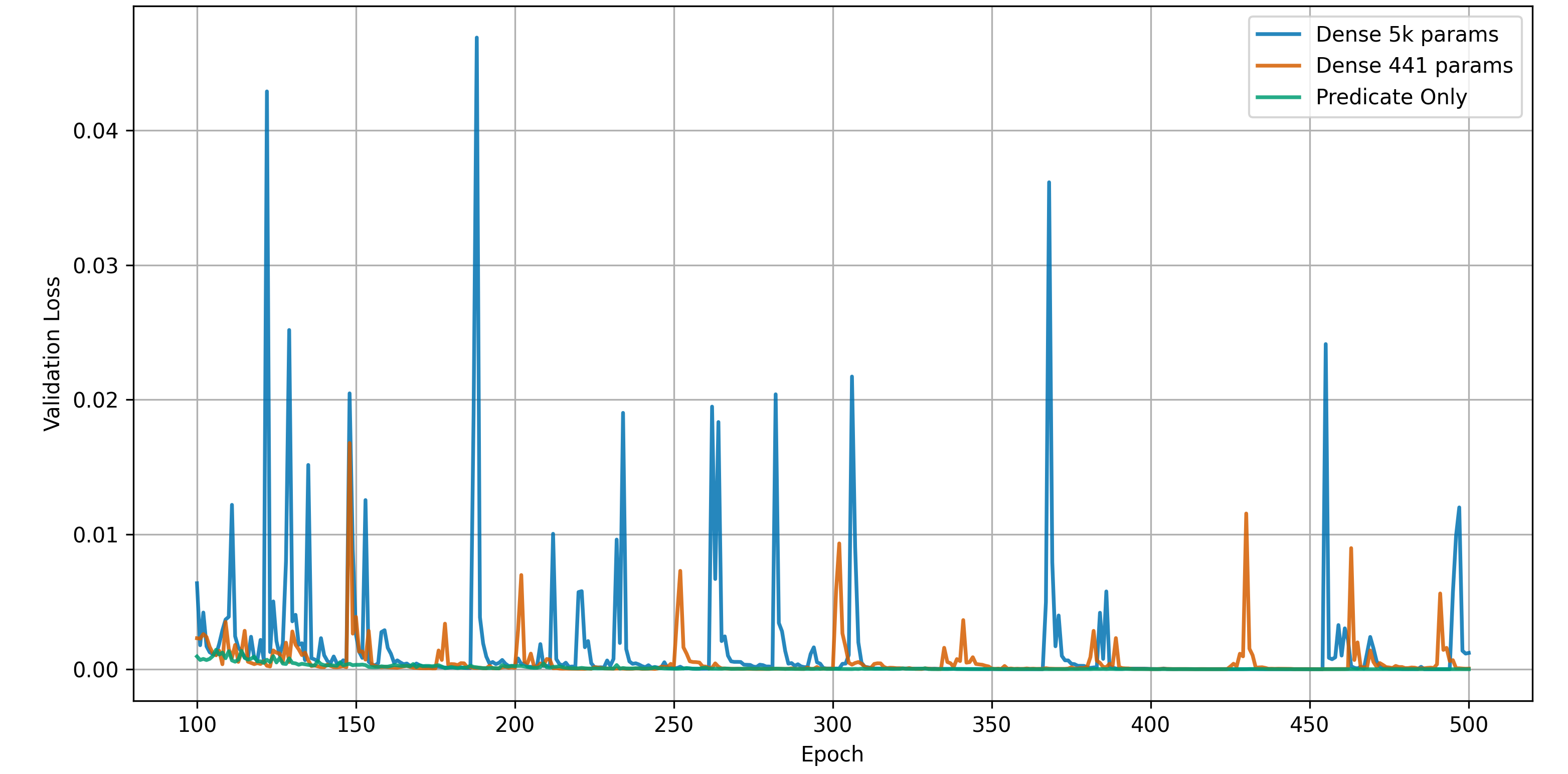}
        \caption{}
    \end{subfigure} \hfill
    \caption{Validation loss of the proposed Logic Network with Predicate Layer (green graph), a Dense network with $441$ trainable parameters (orange graph), and a Dense network with $5k$ parameters (blue graph). The bottom figure depicts the same graph as the top one, but zoomed in.}
    \label{fig:20_atoms_experiment}
\end{figure}

This experiment further proves that the Predicate Neuron extends neural networks with reasoning capabilities, without using Logic Programming auxiliaries to facilitate learning. We observe from the Fig. \ref{fig:20_atoms_experiment} that the proposed neuron speeds up convergence compared to the simple Dense architecture (see graph (a)). Moreover, as can be seen from graph (b) in Fig. \ref{fig:20_atoms_experiment}, both Dense models exhibit sharp artifacts in validation loss, while the graph loss of the proposed Logic Network is almost perfectly flat (close to $0$). Overall, our goal is not to use neural networks to solve/approximate logic reasoning tasks, but to natively embed predicate knowledge in standard architectures and achieve better accuracy and overall effectiveness.

Additionally we benchmark two Neurosymbolic Neural Networks a LTN and a Differentiable LGN. The LTN consists of a $20$-input layer, $2$ Dense layers with $256$ units and ELU activation, as well as an $21$-output layer with Sigmoid activation. As for the LGN network, it includes a $20$-input layer, $8$ Logic Gate Layers (following a "straight" architecture as in \cite{Petersen2022deep}), a Logic Gate layer producing $336$ gates ($21 * 16$, i.e. $16$ separate gate units created for each of the $21$ outputs)), and a layer that aggregates the output neurons, ultimately yielding $21$ outputs, that are bounded in the range of $[0, 1]$ by a Sigmoid activation layer. Table \ref{tab:20_atom_LTN_LGN} contains the test loss of all three models, i.e. LGN, LTN and the Logic Network integrating Predicate Neuron, from which we deduce that the latter model outperforms both Neurosymbolic architectures.

\begin{table}[htbp]
    \centering
    \vspace{0.3cm}

    \begin{threeparttable}
        \begin{tabular}{l|c}
            \hline
       Model & Test Loss \\
       \midrule
       Logic Network  & $9.8694*10^{-7} \pm \sim1*10^{-8}$ \\
       LTN  & $4.5123*10^{-2} \pm 2.7*10^{-4}$ \\
       LGN  & $2.6650*10^{-6} \pm \sim1*10^{-8}$  \\
       \hline
    \end{tabular}
    \caption{Comparison between LTN, LGN and a Logic Network on the $20$-atom problem.}
    \label{tab:20_atom_LTN_LGN}
    \end{threeparttable}

    \vspace{0.3cm}
\end{table}

\begin{figure}[htbp]
    \centering
    \begin{subfigure}[b]{0.90\textwidth}
        \centering
       \includegraphics[width=\textwidth]{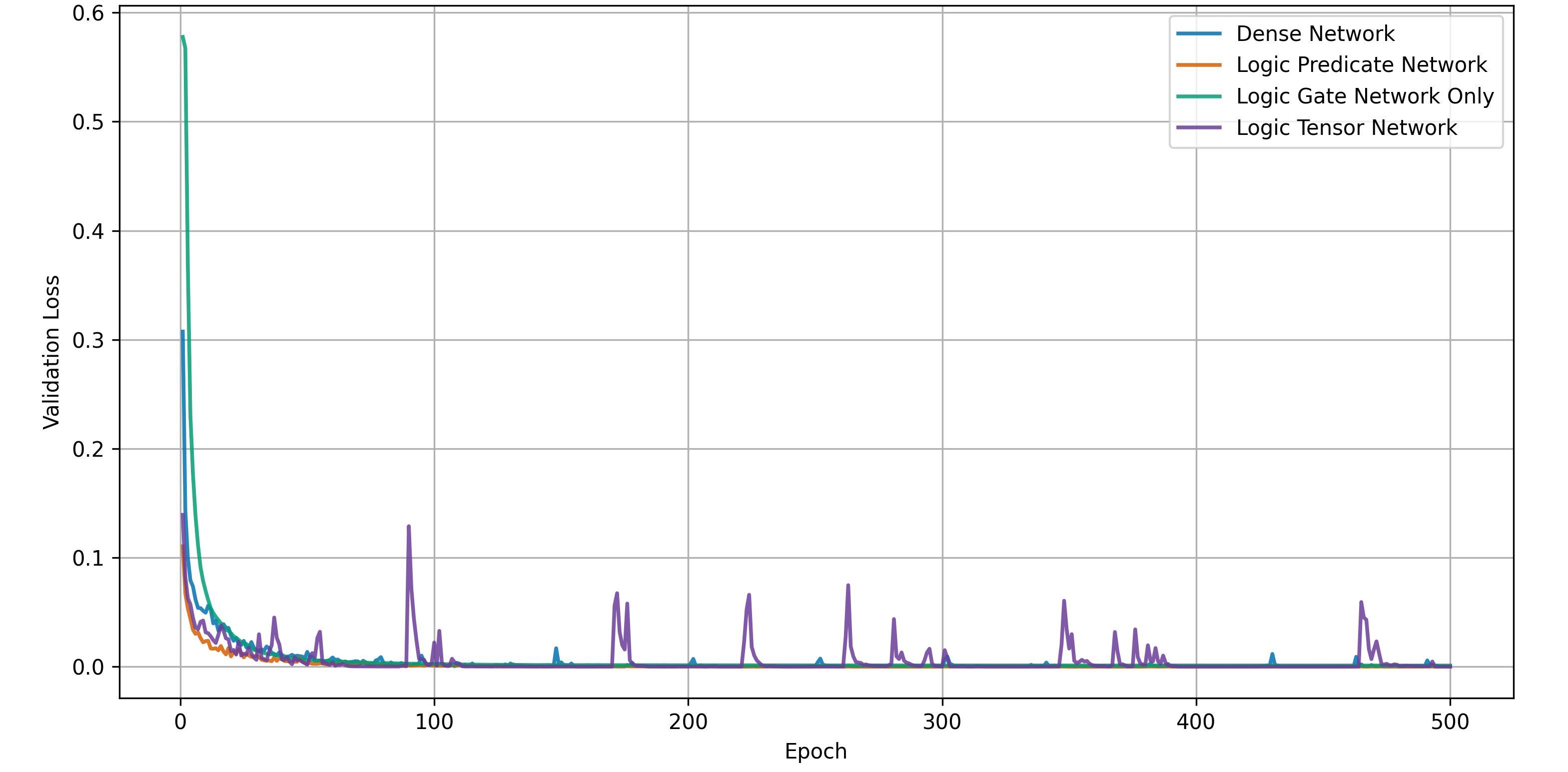}
       \caption{}
    \end{subfigure} \hfill
    \begin{subfigure}[b]{0.90\textwidth}
        \centering
        \includegraphics[width=\textwidth]{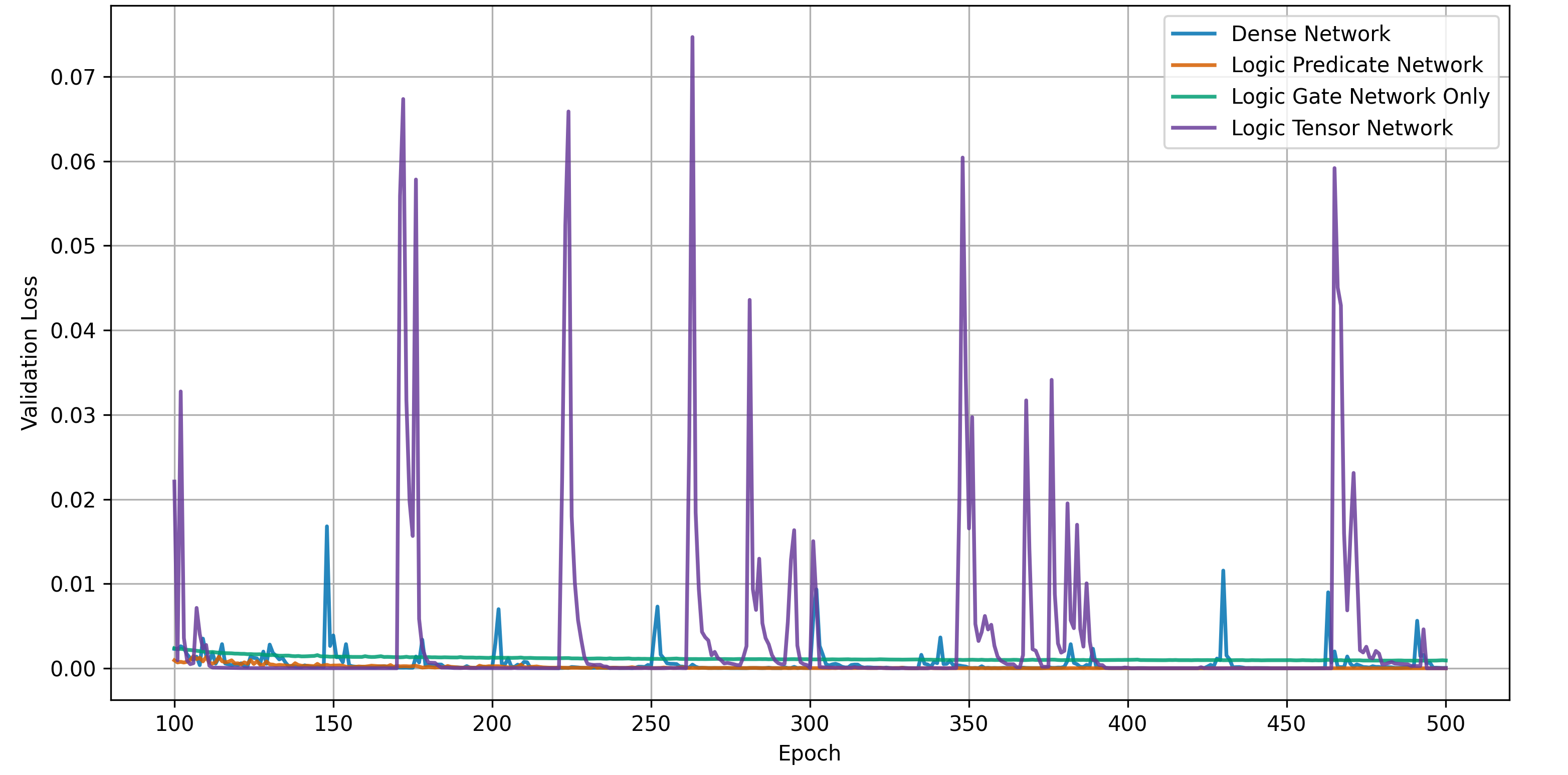}
        \caption{}
    \end{subfigure} \hfill
    \caption{Validation loss of the proposed Logic Network with Predicate Layer (orange graph), a Dense network with $5k$ parameters (blue graph), an LTN network (purple graph) and an LGN network (green graph). The bottom figure depicts the same graph as the top one, but zoomed in.}
    \label{fig:20_atoms_experiment_LGN_LTN}
\end{figure}

Figure \ref{fig:20_atoms_experiment_LGN_LTN} depicts the test loss that the LTN, LGN, Logic Network, and the Dense model with $5k$ parameters achieve. It is evident that the model incorporating Predicate Neurons learns faster reaching a very small loss even from the first $50$ epochs. Both the latter network and the LGN are the most stable ones learning the logic of the problem without exhibiting any loss artifacts, unlike LTN and Dense models as can be also seen in graph (b) of Fig. \ref{fig:20_atoms_experiment_LGN_LTN}. Even though LGN captures the problem, converges slower and cannot further reduce its loss more than an order of $10^{-2}$.

\subsubsection{Experiments on Non-continuous Functions}
\label{sec:non_continuous_experiments}

In this experiment, we intend to show that the Predicate Neuron natively resolves discontinuities by leveraging a two-step quantization strategy: i)  bounding the logic network's input in the range of $[0, 1]$ by passing it through a dense layer with sigmoid activation, ultimately creating discrete fuzzy predicates, ii) quantizing the model's internal logic relation state by employing only one trainable parameter ($\theta$), in the Predicate Neuron. This quantization strategy allows the logic network to overcome the non-continuous nature of the to-be-approximated function. For the purposes of this experiment, we create a dataset based on the branched function defined in Eq. \ref{eq:branced_function}:

\begin{equation}
f(x) =
\begin{cases}
x & x < -200 \\
-x & -200 \le x < 200 \\
2x & x \geq 200
\end{cases}
\label{eq:branced_function}
\end{equation}

Hence, the Logic Network takes as input the values of parameter $x$ and outputs the respective $f(x)$. The logic network consists of one input, a fully-connected layer with $3$ neurons followed by a Sigmoid activation, $3$ Predicate Neurons ($1$ Maximal Predicate Layer with $3$ inputs), one fully-connected layer with $2$ units and sigmoid activation, another $2$-unit dense layer, whose output is multiplied with the input $x$ and is fed to a single-unit Fully-Connected layer to produce the output. We refer to the neural block preceding the single-unit dense layer, as "Logic Core". This architecture illustrates an example of integrating predication in neural networks. Just as in the above two experiments, we compare the Logic Network containing our proposed Predicate Neuron with Fully-Connected architectures, and especially three other models. To make fair comparisons, two of the three Dense models follow the same neuron architecture as the Logic Network, but instead of incorporating Predicate Neurons in their Logic Core, they employ a $10$-unit dense with ReLU and without activation, respectively, while the third Fully-Connected network has a simpler structure with one input, two dense layers with $3$ units and one output.

\begin{figure}[h!]
    \centering
    \includegraphics[width=\textwidth]{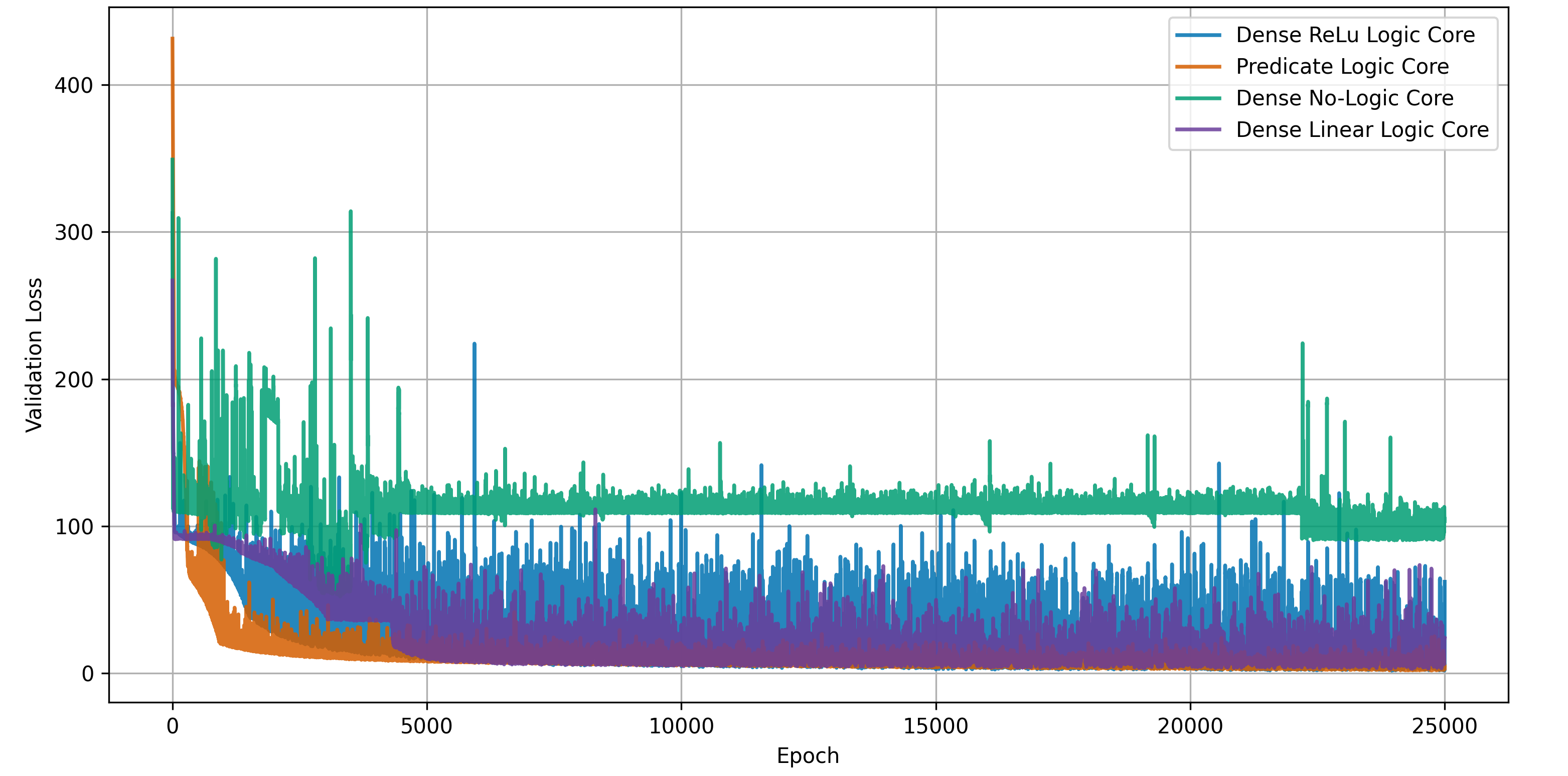}
    \caption{The validation losses across $25000$ epochs for: i) a Dense network without Logic Core (green), ii) a Dense network with Logic Core (purple), iii) a Dense model with Logic Core and ReLU activation in the second layer of its Logic Core (blue), and iv) the Logic Network incorporating Predicate Neurons in their Logic Core (orange).}
    \label{fig:linear_val_losses}
\end{figure}

Figure \ref{fig:linear_val_losses} illustrates the validation loss for the Logic Network (the one enhanced with the Predicate Neuron - with orange color), the simple Dense network (green graph in the same figure), a Dense network with Logic Core (purple graph), and a same Fully-Connected network but with ReLU activation function in the second layer of its Logic Core (blue graph), with the Logic Network converging and exhibiting minimal loss from the first $\sim1000$ epochs. On the contrary, the Dense architecture without deep logic (green graph) produces a large validation error ($\sim100$), while the Dense networks with linear Logic Core (blue and purple graphs) reach a validation loss similar to the one of the Logic Network, but with a slower convergence rate (i.e. the converging at approximately $5000$ epochs). More specifically, the testing accuracy of all models is measured in Mean Absolute Error (MAE), with the simpler Dense network (i.e. green graph in Fig. \ref{fig:linear_val_losses}) not being able to learn discontinuities, achieving a large error on the validation set, $MAE = 93.8606$, while the Logic Network performs notably better with a validation error of $MAE = 1.7830$. This large error difference between these two models is attributed to the fact that standard neural networks incorporating continuous activation functions (e.g. ReLU, Sigmoid, Tanh, etc.) exhibit spectral bias, meaning that they focus on primarily learning smooth, low-frequency functions, while fitting to high-frequency functions or sharp jump discontinuities poses as a challenge for them. Explanatory, a Dense network tries to "smooth out" the abrupt step jumps defined in Eq. \ref{eq:branced_function} rather than learning a step function.

It also is worth mentioning that the Fully-Connected network without Logic Core (green graph in Fig. \ref{fig:linear_val_losses}) seems to converge to zero training loss during learning, but it completely fails in the validation set, which we assume is because it simply overfits on the training dataset, not being able to extract relations that would generalize to other cases.
Regarding training times, the simpler Dense model performs $25000$ in $374.27$ seconds, and the Logic Network along with the Dense networks with Logic Core run the same epochs in almost double the time, with the Predicate Logic Network at $539.63$ seconds.

\subsubsection{The Parity Experiment}
\label{sec:parity_experiment}

To test the limits of the proposed neuron on the range of problems it may approximate effectively, we evaluate the Predicate Neuron on the parity problem. The parity problem refers to the task where the model is fed with $N$ binary inputs, and it has to deduce whether the number of ones included in the inputs is an even or an odd number, a problem that requires capturing higher-order non-linear relations in the training data, with which single-layer perceptrons and Dense networks excluding very deep ones, are not able to learn \cite{Minsky2017perceptrons, Rumelhart19881986}. The logical network employed for this task contains a $20$-input layer, $5$ hidden layers, with $10$, $5$, $3$, $2$, and $1$ Predicate Neurons, respectively, and one output. The latter architecture is called the Minimal Predicate Layer. Subsequently, we delve into a sensitivity analysis in terms of the learning rate required for learning deep logic.

\begin{figure}[htbp]
    \centering
    \begin{subfigure}[b]{0.60\textwidth}
        \centering
       \includegraphics[width=\textwidth]{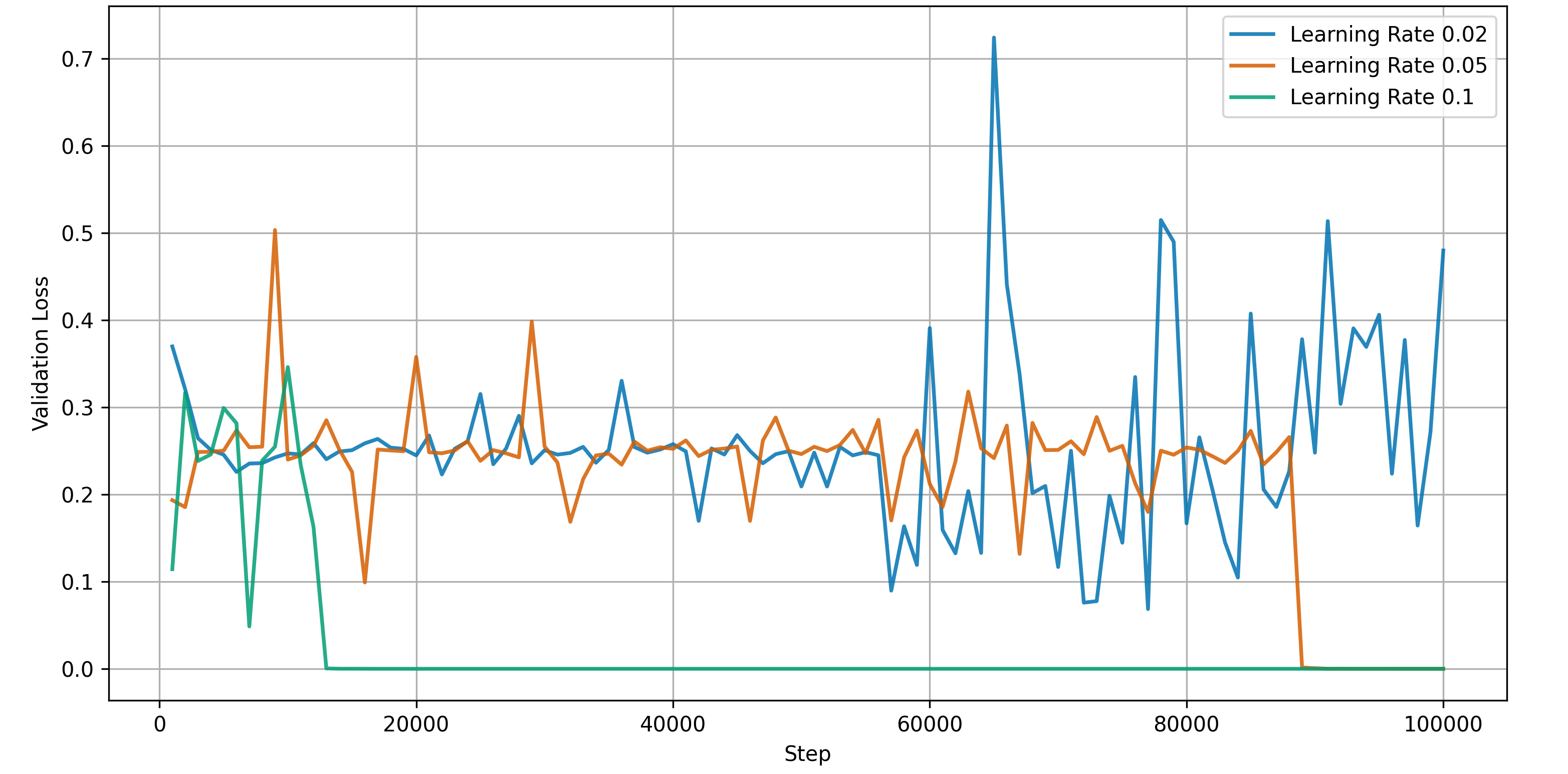}
    \caption{}
    \end{subfigure} \hfill
    \begin{subfigure}[b]{0.60\textwidth}
        \centering
        \includegraphics[width=\textwidth]{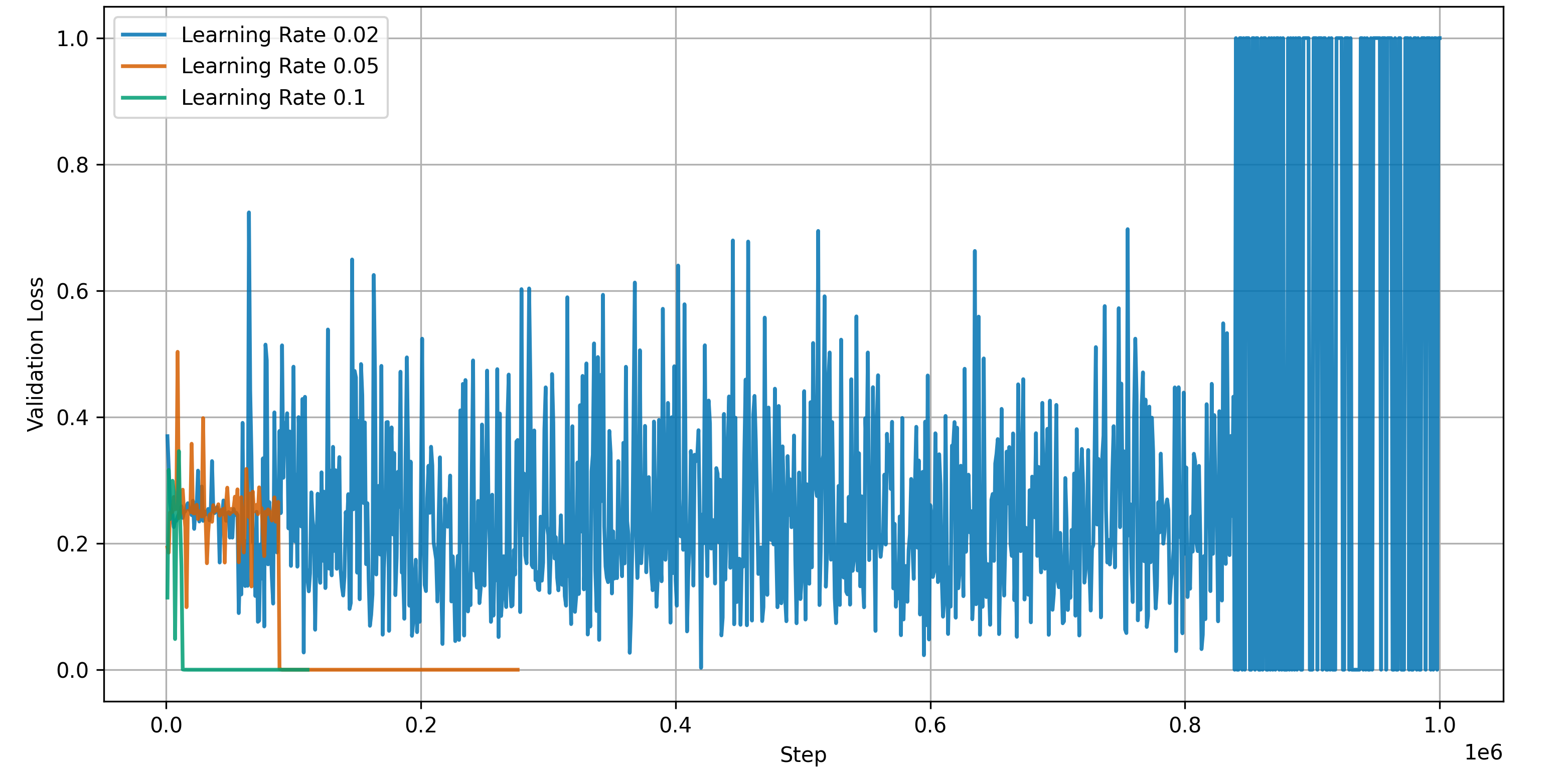}
     \caption{}
    \end{subfigure} \hfill
    \begin{subfigure}[b]{0.60\textwidth}
        \centering
        \includegraphics[width=\textwidth]{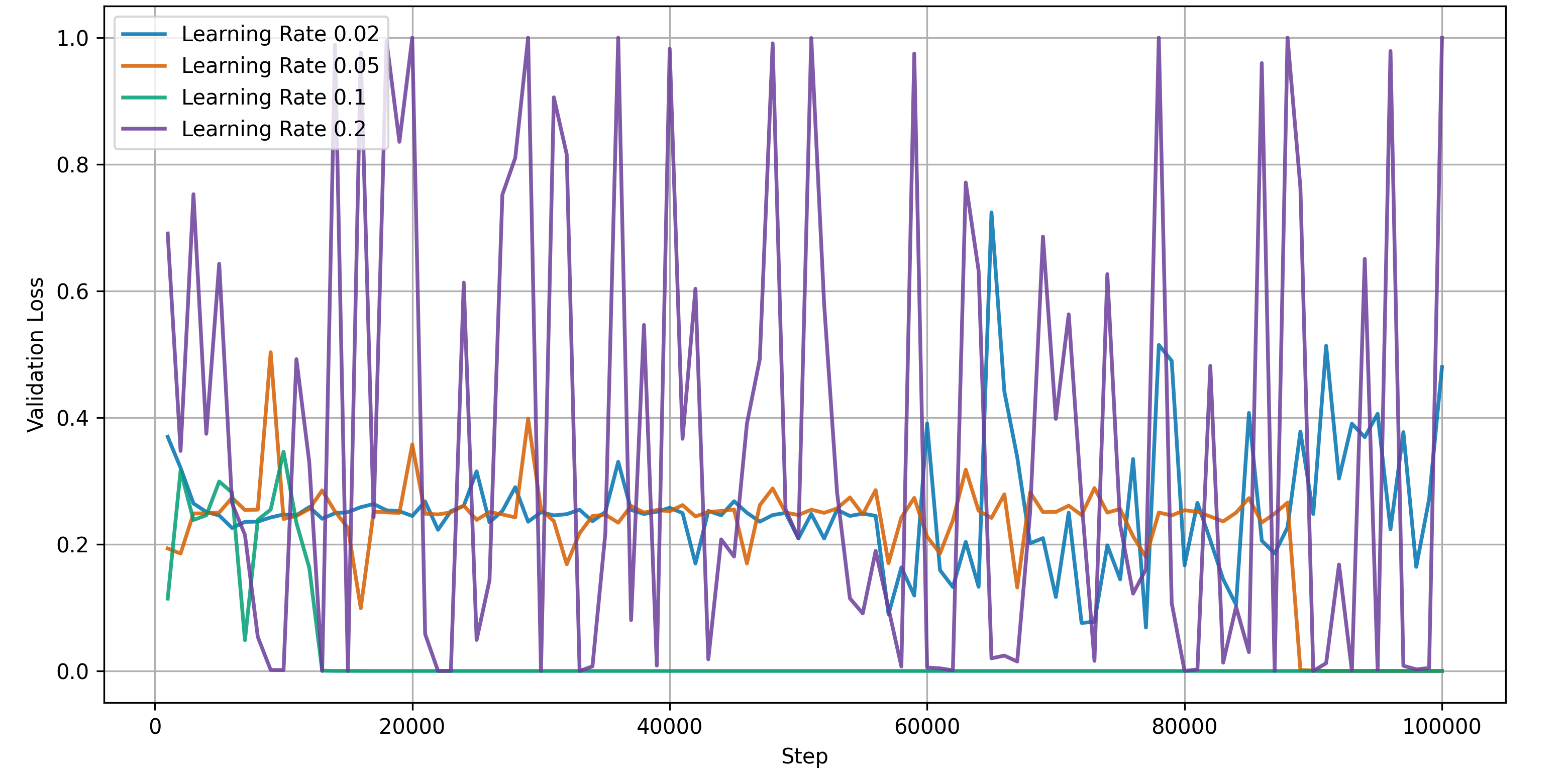}
        \caption{}
    \end{subfigure} \hfill
        \begin{subfigure}[b]{0.60\textwidth}
        \centering
        \includegraphics[width=\textwidth]{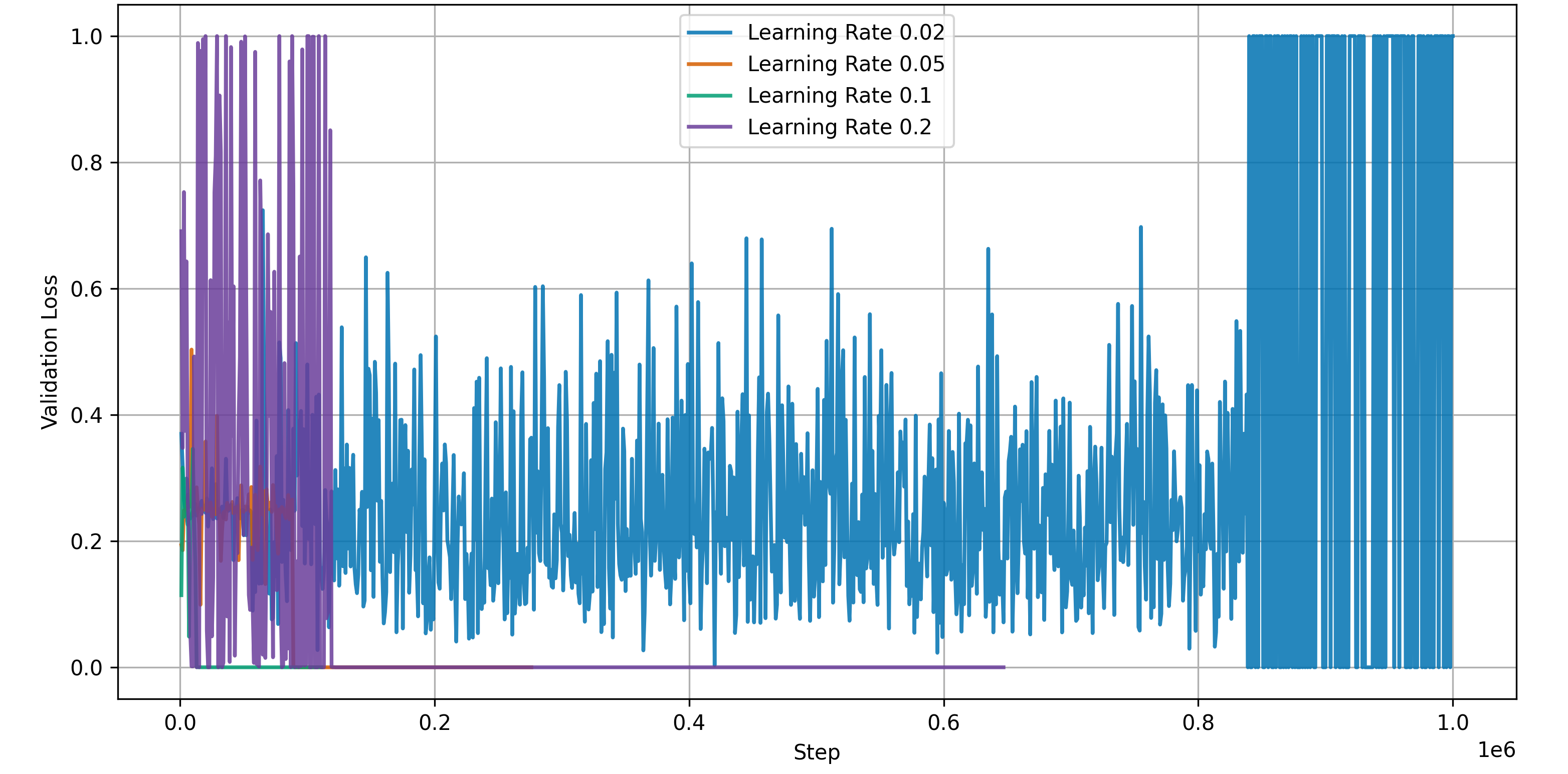}
       \caption{}
    \end{subfigure} \hfill
    \caption{Training loss of the Logic Network across different learning rate values. Figures (c) and (d) are the same as (a) and (b), but also depict one extra experiment (purple graph). The training loss is illustrated per-step, since usually the network converges within one epoch.}
    \label{fig:parity_experiment}
\end{figure}

Figure \ref{fig:parity_experiment} shows a very interesting finding; the larger the learning rate, with an upper soft bound close to $0.1$, the faster the model converges within an epoch, ultimately learning the underlying logic of the problem. This is a counter-synergistic behaviour that comes in contrast with conventional neural networks that converge easier usually using small learning rates. Specifically, an adequately large learning rate of $0.1$ converges in $\sim15000$ steps, with the optimizer exploring neighboring values (green graph in Fig. \ref{fig:parity_experiment}). Smaller learning rates, such as $lr = 0.02$ and $lr = 0.05$ (graphs with blue and orange color respectively in Fig. \ref{fig:parity_experiment}) eventually reach convergence after many oscillations due to the optimizer experimenting with possible solutions that are very distant with one another, i.e. basically jumping between binary values producing large errors. Yet, once again the larger learning rate, the one of $lr = 0.05$ (orange graph) aids the model to approximate a preferred solution earlier, i.e., in $\sim90000$ steps, than the smaller one, while the $lr=0.02$ guides the network to instability causing high oscillations towards the last steps of the epoch as shown in (b) in Fig. \ref{fig:parity_experiment}. Learning rate of $0.2$ (purple graph) seems to produce the same behavior as the smaller learning rates during the first $\sim150000$ steps of the epoch, but quickly reaches convergence, in contrast to when employing smaller learning rates. It is also worth mentioning that, during training, when the training loss is smaller than $0.1$, then the soft forward pass switches back to the hard forward pass, since the model has already captured the logic of the problem. The latter causes the training loss reaching an absolute $0$, as can be seen in Fig. \ref{fig:parity_experiment}, consolidating the model's convergence to an accurate solution.

For consistency, we performed experiments with many LTN and LGN architectures on the parity problem. Regarding LGNs, we also evaluated architectures with the same reducing width per layer structure that the Minimal Predicate has. In one of our experiments, we devised such an LGN structure initialized solely with XOR gates, which achieved $0$ loss and produced always accurate results, that of course is not trained as we supplied the solution as the initial condition to the network. The latter outcome indicates that the LGN has the capacity to express the solution of the parity problem, though it seems that through gradient-based learning is failing to converge to the correct gate configuration, with any learning rate. A notable result of all the experiments LTNs and LGNs for approximating the parity problem is that architectures exhibiting the lowest error seem to stabilize at exactly $MSE = 0.25$, which is $0.5^2$ with $0.5$ being the mean between the two possible output values, $0$ and $1$.

For the purpose of this experiment, we introduce a more complex architecture incorporating both Dense and Predicate Neurons. The first layer of this model has $3$ neurons, where the first one has trainable parameters $w = 0.081$ and $b = 12.99$. The neuron converges to such values for weight and bias, since when $x < -200$, i.e. $wx + b = 0.081 *(-200) + 12.99 = -16.3 + 12.99 = -3$, then the sigmoid results to a value close to $0$. For this neuron's result to be closer to $1$ then $x$ has to have a much higher value than $-200$. So this neuron can be said to approximate the Predicate $Y_a$ = "x is larger than -200", in a soft, rather than perfect, sense, consistent with fuzzy logic. The second neuron has $w = 0.066$, and $b = -11.6$. In the same sense as the previous one captures in a soft sense the predicate $Y_b$ = "x is less 200". As for the third neuron, it has $w = 0.85$, $b = 0.53$ so we can say that it captures the predicate $Y_c$ = "x is larger than 0". The second layer of this model is the Predicate Layer, a layer where all $3$ previous predicates are pairwise fully-connected to Predicate Neurons. The first Predicate Neuron has $\theta = 3.346$ (XOR) between $Y_a$ and $Y_b$, which results in the predicate "x is below -200 or above 200". The second Predicate Neuron has $\theta = -1.375$ (AND) between $Y_a$ and $Y_c$, which results in the predicate "x is above 0". The third Predicate Neuron has $\theta = -1.372$ (AND) between $Y_b$ and $Y_c$, which results in the predicate "x is above 0 and below 200". These $3$ predicates are enough to convey all the possible states of the branched function if AND or NOT operation is performed between two of them, e.g. "x is above 0" AND "x is below -200 or above 200" means "x is above 200". These predicates are used by the Dense network thereafter to output the result based on which is true.

\subsection{Benchmarks on Typical Architectures}
\label{sec:bench_typical_architectures}

The above four experiments outlined in Section \ref{sec:fund_benchmarking} were focused on whether the Predicate Neuron will aid Fully-Connected architectures to learn faster, more effectively, and accurately. In the set of experiments outlined in this and the subsequent Sections, we integrate the Predicate Neuron in neural structures other than Dense ones, starting with the standard Convolutional Neural Network (CNN), and exploring also Recurrent and Transformer-based models.

\subsubsection{Integration with Convolutional Neural Networks}

We proceed with evaluating the Predicate Neuron within a CNN model for a classification task, for which we devise a CNN architecture with $2$ convolutional layers, one with $32$ filters and one with $64$ filters, both with a kernel size of $3$, and MaxPooling layers succeeding each one of them. The network's classifier consists of a dense layer with $128$ units, a dropout layer and a fully-connected output layer with $10$ units. The logic network employed for this experiment follows the same CNN architecture, but we also incorporate a logic core comprising of a $10$-unit dense layer with sigmoid activation, a Maximal Predicate Layer (as described in \ref{sec:20_atom_experiment}), and a dense layer with $10$ units, between its convolutional and classifier parts. The same quantization strategy described in Section \ref{sec:non_continuous_experiments}, is also applied through the aforementioned logic core. The output of the logic core is passed through a Lambda layer that shifts its values by $0.5$. There are many ways to apply logic as signal within the neural network, in this example the Lambda layer serves as a multiplier from $0.5$ to $1.5$, i.e. weakening or strengthening signals that it is later multiplied with. The output of the Logic Core and the one of the classifier are multiplied yielding the output of the Logic network. We, then, compare the CNN model with (named Logic CNN) and without employing the Logic Core alongside it, based on the classic MNIST dataset \cite{Lecun1998gradient}. Both models were evaluated using the Sparse Categorical Cross-entropy loss and were trained for $25$ epochs, on the validation set containing $20000$ images.

Table \ref{tab:MNIST_results} summarizes our findings. We deduce that adding the proposed neuron into the CNN architecture increases the testing accuracy by only a small percentage of $>0.01\%$, however the test loss is reduced by $11.3\%$. The training time of the logic network indicates that it converges faster than the plain CNN model.

\begin{table}[htbp]
    \centering
    \vspace{0.3cm}

    \begin{threeparttable}
        \begin{tabular}{l|c|c|c|c}
            \hline
       Model & Parameters & Training Time (Secs) & Test Loss & Test Accuracy \\
       \midrule
       Logic CNN Network  & $241549$ & $141.1219$  & $0.03280 \pm 0.00053$ & $0.9930 \pm 0.00119$ \\
       Baseline CNN  & $225034$ & $143.5178$ & $0.03651 \pm 0.00061$ & $0.9929 \pm 0.00130$ \\
       \hline
    \end{tabular}
    \caption{Results on MNIST dataset.}
    \label{tab:MNIST_results}
    \end{threeparttable}

    \vspace{0.3cm}
\end{table}

\subsubsection{Integration with RNN architectures}

The Predicate Neuron is also assessed alongside recurrent architectures on the Permuted MNIST \cite{Kirkpatrick2017}, a dataset based on the original MNIST images with pseudo-random pixel permutation, usually used for regression tasks by modifying the class target labels to continuous values. In the context of this evaluation, we integrate the Minimal Predicate Layer (see subsection \ref{sec:parity_experiment}) into an LSTM-based network. In particular, the Logic Network comprises of an LSTM layer with $128$ units, whose output is normalized in the $[0,1]$ range via a Lambda layer and fed to a Minimal Predicate Layer. The elements of the latter layer's tensor are interleaved along the column axis to match the subsequent $128$-unit fully-connected layer, whose output is multiplied by the output of the initial LSTM layer. This model is compared against an LSTM network, containing a $128$-unit LSTM layer and a Dense $10$-unit output layer, essentially a simple RNN model without a Logic Core. Both models were trained for $100$ epochs, using a $128$ batch size and a learning rate of $0.001$.

\begin{figure}[h!]
    \centering
    \includegraphics[width=\textwidth]{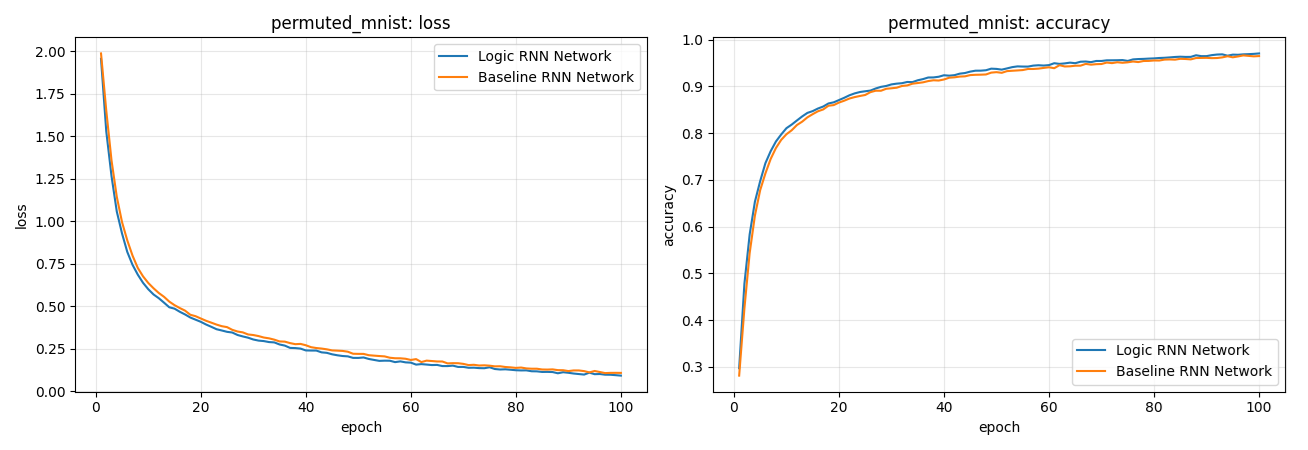}
    \caption{Validation loss and accuracy of the RNN network with Predicate Layer (blue graph), and a Baseline RNN without a Logic Core (orange graph).}
    \label{fig:rnn_loss_accuracy}
\end{figure}

Our results are summarized in Table \ref{tab:rnn_results}, from which we observe that the RNN with Logic Core performs better in terms of testing accuracy with an improvement of $\sim0.8\%$, while test loss has reduced by $5.5\%$. Furthermore, the Logic RNN network trains $3.85\%$ faster than the Baseline RNN model, even though the former has $24.43\%$ more parameters. The testing loss and accuracy of the Logic RNN and Baseline RNN models are depicted in Fig. \ref{fig:rnn_loss_accuracy}.

\begin{table}[htbp]
    \centering
    \vspace{0.3cm}

    \begin{threeparttable}
        \begin{tabular}{l|c|c|c|c}
            \hline
       Model & Parameters & Training Time (Secs) & Test Loss & Test Accuracy \\
       \midrule
       Logic RNN Network & $84427$ & $25151.5$  & $0.3578 \pm 0.0057$ & $0.9115 \pm 0.0046$ \\
       Baseline RNN & $67850$ & $26119.5$ & $ 0.3777 \pm 0.0071$ & $0.9043 \pm 0.0053$ \\
       \hline
    \end{tabular}
    \caption{Results on Permuted MNIST.}
    \label{tab:rnn_results}
    \end{threeparttable}

    \vspace{0.3cm}
\end{table}

\subsubsection{Integration with Transformers}

\begin{figure}[h!]
    \centering
    \includegraphics[width=0.80\textwidth]{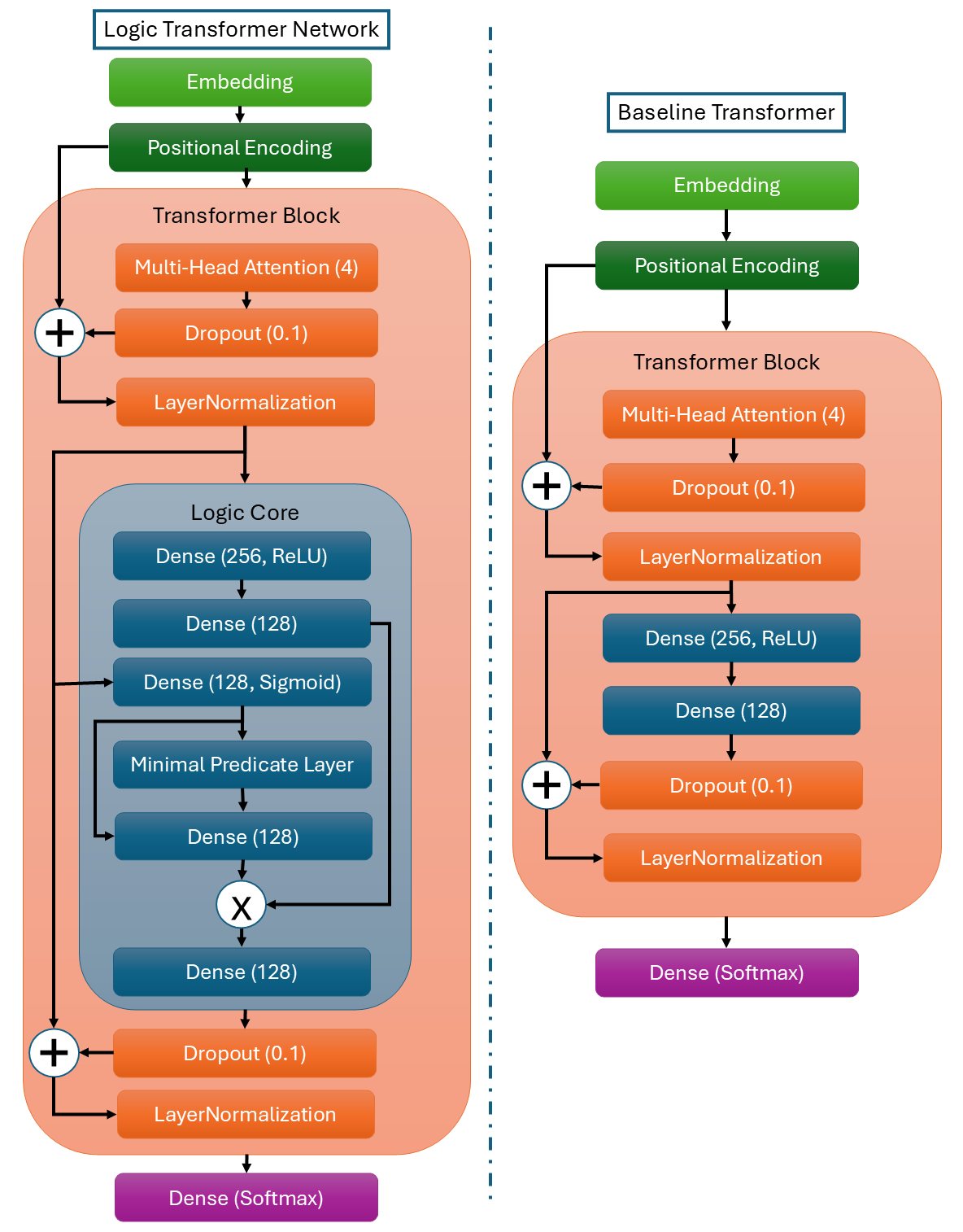}
    \caption{The architectures of the Baseline and Logic Transformer }
    \label{fig:transformer_architectures}
\end{figure}

We also incorporate Predicate Neuron Layers in Transformer models employed for next-token prediction. Following the previous experiments, we design two Transformer-based architectures, one incorporating a Logic Core and one without. As illustrated in Fig. \ref{fig:transformer_architectures}, both models consist of an Embedding and a Positional Encoding layer, a Transformer-based module, and an output Fully-Connected layer with softmax activation. The Transformer-based module for the simple architecture contains a Multi-head Attention layer with $4$ heads followed by a Dropout layer and a Normalization layer, $2$ Dense layers, one with $256$ units and ReLU activation, and one with $128$ units whose output is passed through a Dropout, and a Normalization layer as well. The Transformer module for the Logic Network follows the same structure, yet we substitute the $2$ Fully-Connected layers with a Logic Core incorporating a Dense $256$-unit layer with ReLU activation, a $128$-unit Dense layer, and another $128$-unit Dense layer with Sigmoid activation producing fuzzy predicates for the subsequent Minimal Predicate Layer. The output of the Predicate Layer is fed to a $128$-unit Dense layer and the produced tensor is multiplied with the output of the first $128$-unit Fully-Connected layer of the Logic Core, with the result of this multiplication passing through another Dense $128$-unit layer.

We train both the proposed Transformer-based models on the Wikitext-103 dataset \cite{Merity2016, Merity2018}, a $ 110$-times larger subset of the Wikitext \cite{Merity2016} original data, containing over $100$ million tokens from untruncated verified Good and Featured English articles on Wikipedia, tailored for long-term dependency language modeling, thus posing as a standard benchmark for Transformer and RNN language models. Before employing this dataset for training and testing the Transformer-based networks, we convert texts to token sequences, based on a constructed simple word-level token vocabulary of size $ 10000$. We train our models for $10$ epochs, with a learning rate of $0.001$ and a batch size of $16$, while using a Sparse Categorical Cross-entropy loss.

\begin{table}[htbp]
    \centering
    \vspace{0.3cm}

    \begin{threeparttable}
        \begin{tabular}{l|c|c|c|c}
            \hline
       Model & Parameters & Training Time (Secs) & Loss & Accuracy \\
       \midrule
       Logic Transformer Network & $2.917.776$ & $3134$  & $3.8187 \pm 0.0344$ & $0.5335 \pm 0.0036$ \\
       Baseline Transformer & $2.834.960$ & $2479$ & $3.8685 \pm 0.0889$ & $0.5261 \pm 0.0058$ \\
       \hline
    \end{tabular}
    \caption{Results on Wikitext-103.}
    \label{tab:transformer_results}
    \end{threeparttable}

    \vspace{0.3cm}
\end{table}

From Table \ref{tab:transformer_results}, we deduce that the Logic Transformer Network exhibits a loss of $3.8187$, meaning a $1.3\%$ reduction in loss after solely $10$ epochs, compared to the simple Transformer architecture (loss of $3.8685$), and a respective $1.4\%$ improvement in accuracy. It is worth noting that the model is learning a meaningful language structure, reaching $\sim53\%$ top-5 accuracy, meaning that approximately $53\%$ of the time, the correct next token appears among the model's $5$ most likely predictions. Accuracy is not the most insightful way to evaluate a language model, yet the perplexity metric is usually used, which is formulated as shown in Eq. \ref{eq:perplexity} \cite{Takahashi2019}:

\begin{equation}
Perplexity = e^{-\frac{1}{N} \sum_{i=1}^{N}logq(x_i)}
\label{eq:perplexity}
\end{equation}

where $\sum{i=1}^{N}logq(x_i)$ is the cross-entropy loss measuring the model's uncertainty in predicting the next token (i.e. $q(x_i)$ the predicted probability of the word $i$ being present in the given text), ultimately yielding $e^{loss}$. Given said that and taking into consideration these simple Transformer models, both architectures exhibit a good perplexity, $\sim47.8705$ for the Transformer model without Logic Core and $\sim45.5449$ for the Logic Transformer (i.e. $5.1061$ improvement in perplexity), with WikiText-103 Transformer and RNN benchmarks having a perplexity of approximately $18-40$ \cite{Dai2019transformer, Merity2018analysis, Grave2016improving}. However, in the context of this paper, we do not aim to design a high-end language model, but to showcase whether the Predicate Neuron Layers can improve the performance of Transformer-based architectures.

\section{Discussion - Conclusion}

In this work, we propose the Native Predicate Neuron, a differentiable logic reasoning neural module that bestows typical neural architectures with reasoning capabilities, ultimately enhancing testing accuracy and convergence. The neuron models fuzzy logical operations, and learns to dynamically switch between them, leveraging a single trainable parameter. During training, we substitute the hard output of the forward pass with a smooth formulation to enable differentiability during the backward pass, via a tailored surrogate gradient mechanism. Furthermore, we introduce two custom layers, namely the Maximal and the Minimal Predicate Layers, comprising entirely of Predicate Neurons, which can be easily integrated into standard deep learning models. A quantization strategy to quantize both the Predicate layer's input and internal state is employed to facilitate convergence in case of approximating non-continuous operations, without requiring to explicitly handle discontinuities.

We conducted two sets of experiments to evaluate the effectiveness of the proposed neuron: i) we benchmarked the Predicate Neuron in simple logic problems and more complex ones, while comparing against Fully-Connected neural structures, and ii) we embedded Predicate layers in typical deep learning architectures including CNNs, RNNs, and Transformer models. At this point it is worth mentioning, that the evaluations carried out in Section \ref{sec:bench_typical_architectures} were conducted upon prominent datasets that are not tailor-made for the proposed Native Predicate Neuron, but, on the contrary, they are common generic cases for machine learning benchmarking. More specifically, these benchmarks describe typical generic deep learning problems, the approximation of which is improved (even by a small percentage) with neural architectures incorporating the Predicate Neuron. This is a very promising finding, indicating that, in the general case, the addition of a Logic Core does not hinder the performance of the model it was integrated with, whereas it ameliorates it. In specific cases, where logic relations are part of the nature of the problem or dataset, the Predicate Neuron is capable of capturing features and logic relations within a dataset, which a neural network without Logic Core, would struggle with or could not even capture at all.

Our results illustrate that Predicate Layers, as well as Dense architectures incorporating Predicate Neurons, outperform the respective Fully-Connected models in terms of testing loss, while also converging earlier, even $x5$ times less training steps as shown in Subsection \ref{sec:non_continuous_experiments}, during training, compared to their standard, without logic core, Dense counterparts. Moreover, CNN and LSTM-based architectures with Logic Cores, i.e., neural architectures integrating Predicate Neurons/Layers, exhibit a small improvement in testing accuracy, but a non-negligible decrease in testing loss on MNIST (i.e. $11.3\%$ reduction) and permuted MNIST (i.e. $5.5\%$ decrease), respectively, indicating that the incorporation of Predicate Modules can positively alter the models' effectiveness. On the contrary, the Transformer model enriched with Logic Core produces significantly better perplexity, test loss and accuracy results on the Wikitext-103 dataset, unlike the same Transformer architecture without Predicate Neurons. Nevertheless, it is safe to conclude that the inclusion of Predicate Neurons/Layers in neural structures result in improvement in the overall model's accuracy and efficiency.


As for training times, some of the networks integrating Predicate Layers as Logic Cores, are slightly slower than without Logic Cores, since they usually introduce more trainable parameters, except for comparing a single Predicate Neuron against a Dense model resembling its structural complexity, as in the Experiment described in subsection \ref{sec:fund_ops}, where the Predicate model is clearly faster. Within the case of comparing RelU or Linear Logic Cores with Predicate Neuron logic cores, the Predicate Neuron ones are faster, still, the speed loss that including a Logic Core represents is a limitation of the proposed neuron. Moreover, the fact that the Predicate Neuron through the use of a Logic Core improves CNN and RNN structures by a negligible percentage, poses as another limitation. We assume that fusing Predicate Neurons with convolutional or recurrent kernels may induce larger performance improvement, which is proposed as a possible future research direction. This demands of a quantization of every output of any neuron to predicate scale values such that Predicate Neurons can be natively integrated deeper in the fold of a neural network.


As future work, we intend to address explainability in Neurosymbolic AI, constituting a significant research gap, since it is being less addressed in literature \cite{Colelough2025neuro} despite being a major challenge and an indication of trustworthiness and reliability. Models explicitly embedding symbolic relations inherently offer transparency capabilities (i.e. explainability by design \cite{Calegari2020integration}), since they constrain the path towards the correct prediction, which, however, do not provide sufficient insights into the internal processing of neural networks \cite{Zhang2024neuro}. Considering that the Predicate Neuron transforms soft linear relations into hard fuzzy predicates, we may claim that it naturally increases the explainability level of the overall neural structure that it belongs to. Nevertheless, respective experiments to objectively measure the explainability of our proposed neuron as well as incorporating Explainable AI (XAI) models to make it more transparent, are yet to be conducted. In addition, we would like to explore different Predicate Neuron designs, such as modeling more than one logical operation simultaneously, instead of being constrained to the choice of one operation. The latter may allow the Predicate neuron to be applied in more complex logic reasoning problems than the $20$-atom one outlined in Subsection \ref{sec:20_atom_experiment}.

\printbibliography

\end{document}